\documentclass[journal]{IEEEtran}

\usepackage[T1]{fontenc}
\usepackage[utf8]{inputenc}
\usepackage{amsmath,amssymb,amsfonts}
\usepackage{graphicx}
\usepackage{subcaption}
\usepackage{dblfloatfix}
\usepackage{booktabs}
\usepackage[table]{xcolor}
\usepackage{algorithm}
\usepackage{algpseudocode}
\usepackage{array}
\usepackage{float}
\usepackage{url}
\usepackage[hidelinks]{hyperref}

\newcommand{\generals}{\mbox{\textsc{Generals.io}}}

\begin{document}

\title{Superhuman AI for Generals.io \\ Using Self-Play Reinforcement Learning}

\author{Matej~Straka\textsuperscript{1}, Viliam~Lis\'{y}\textsuperscript{2}, and Martin~Schmid\textsuperscript{1,3}\\[3pt]
{\small \textsuperscript{1}Department of Applied Mathematics, Charles University, Prague, Czech Republic\\
\textsuperscript{2}Department of Computer Science, Faculty of Electrical Engineering, Czech Technical University in Prague, Czech Republic\\
\textsuperscript{3}EquiLibre Technologies, Inc.\\
Correspondence: \texttt{straka@kam.mff.cuni.cz}}%
\thanks{This work has been submitted to the IEEE for possible publication. Copyright may be transferred without notice, after which this version may no longer be accessible.}}

\markboth{}{}

\maketitle

\begin{abstract}
We present a superhuman AI agent for \generals{}, a real-time strategy game that combines long-horizon planning, imperfect information, and combinatorially large state and action spaces. Much prior work on large games relies on heavy machinery such as leagues of past opponents, reward shaping, and human data. We instead train a transformer policy with plain policy-gradient self-play on the sparse win/loss reward, and carefully ablate which ingredients carry the performance: an exponential moving average of the parameters is consistently the stronger deployment policy, and training only on the highest-advantage samples halves the cost of a training iteration. With seven days on four GPUs the agent reaches superhuman play: it ranked first in all ten weekly standings of a two-month deployment on the public 1v1 ladder, and it beat the two strongest humans $189$--$67$ and $32$--$18$. We also release a JAX-native simulator that reaches over $40$ million frames per second on one GPU.
\end{abstract}

\begin{IEEEkeywords}
Real-time strategy games, reinforcement learning, self-play, imperfect information, superhuman AI.
\end{IEEEkeywords}

\section{Introduction}
\label{sec:intro}

\IEEEPARstart{G}{ames} have driven much of modern deep reinforcement
learning, from Chess and Go~\cite{alphazero} to
Poker~\cite{deepstack,libratus}, Stratego~\cite{stratego},
StarCraft~II~\cite{alphastar}, and Dota~2~\cite{dota2}.

The training pipeline behind many of these results decomposes into
two nested loops. An \emph{inner loop} trains a single policy with
reinforcement learning---typically a policy gradient---against a
pool of opponents. An \emph{outer loop} constructs that pool: it
decides which policies enter it, how they are weighted, and when to
introduce new ones. Fictitious play and its neural counterpart
NFSP~\cite{nfsp} aggregate past best-responses into an average
policy that becomes the inner-loop opponent; the
double-oracle family~\cite{mcmahan2003double} and PSRO~\cite{psro}
grow a population by best-responding to a meta-Nash over the
existing set; the AlphaStar league~\cite{alphastar} maintains a
structured pool of main agents, main exploiters, and league
exploiters. The outer loop is what differs across approaches; the
inner loop is, modulo neural architecture and regularization, a
policy improvement step. In plain self-play the pool contains only one
policy, the current one, and the outer loop is trivial.

A natural question is whether the outer loop is doing essential
work at scale, or whether a sufficiently well-tuned inner loop
suffices on its own. On standard imperfect-information benchmarks,
Rudolph et al.~\cite{rudolph2025reevaluating} report that a generic
policy gradient with a simple regularizer matches or exceeds full
PSRO, NFSP, and CFR pipelines once carefully tuned. Sokota et
al.~\cite{sokota2025superhuman} extend this to a large game,
reaching superhuman play in Stratego with a single regularized
policy-gradient learner~\cite{sokota2023unified}. At real-time strategy
(RTS) scale, \mbox{OpenAI Five}~\cite{dota2} reached superhuman performance in
Dota~2 using PPO without human demonstrations, but combined
current-policy self-play with an adaptively sampled pool of historical
opponents and dense hand-designed rewards. Here we investigate whether
current-policy-only self-play can achieve top-level performance in
\generals{} using sparse terminal rewards.

We answer this in the affirmative for \generals{}, an RTS with an
active community of several thousand players and
competitive bots. Unlike StarCraft~II or Dota~2, it is light and
simple enough to train on a single GPU while retaining a rich set
of strategic and tactical challenges: resource allocation,
opponent modeling, and deception under fog of war. One rule set
covers three modes (1v1, 2v2, and free-for-all) spanning
two-player zero-sum, team, and $n$-player general-sum play. To make
this kind of comparison cheap and reproducible in the first place,
we also release a JAX-native simulator that reaches tens of
millions of frames per second on a single GPU and exposes all three
modes through a shared interface.

\begin{figure*}[t]
    \centering
    \setlength{\fboxsep}{0pt}\setlength{\fboxrule}{1pt}%
    \begin{subfigure}{0.32\textwidth}
        \centering
        \fbox{\includegraphics[width=\linewidth]{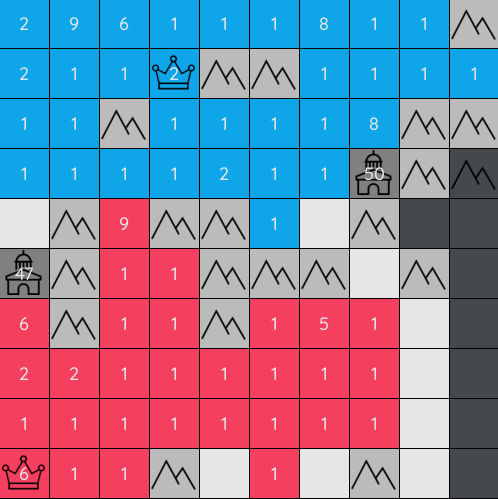}}
        \caption{1v1}
        \label{fig:mode_1v1}
    \end{subfigure}\hfill
    \begin{subfigure}{0.32\textwidth}
        \centering
        \fbox{\includegraphics[width=\linewidth]{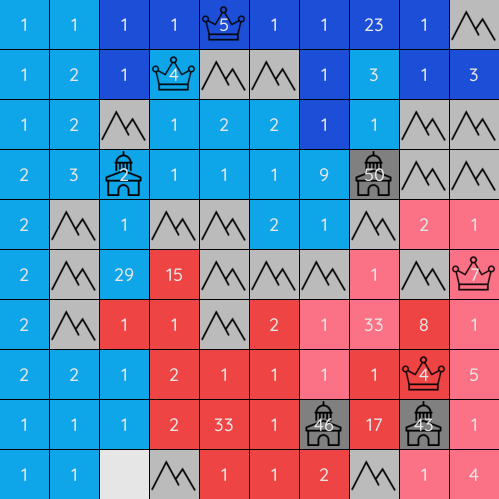}}
        \caption{2v2}
        \label{fig:mode_2v2}
    \end{subfigure}\hfill
    \begin{subfigure}{0.32\textwidth}
        \centering
        \fbox{\includegraphics[width=\linewidth]{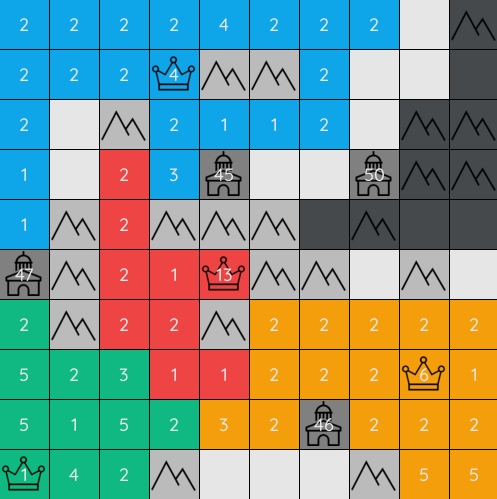}}
        \caption{Free-for-all}
        \label{fig:mode_ffa}
    \end{subfigure}
    \caption{The three game formats supported by the environment.}
    \label{fig:game_modes}
\end{figure*}

\subsection*{Contributions}
\begin{itemize}
\item \textbf{A superhuman agent from policy-gradient
methods}\footnote{The training code is released as open source; available
at \url{https://github.com/strakam/AverageJoe}.} (Sections~\ref{sec:agent} and~\ref{sec:eval}). We train a
transformer policy end-to-end with sparse win/loss reward,
sample filtering, and an exponential moving average (EMA) of its
parameters. The agent ranks first in ten consecutive official weekly
standings of the public 1v1 ladder over a two-month deployment, first
when Elo and OpenSkill ratings are recomputed over all $173{,}776$
ranked games of that period, and wins $82.5\%$ of its $1{,}147$ rated
games, $70\%$ against the players rated $100$ and above.

\item \textbf{Ablations} (Section~\ref{sec:ablations}) that isolate
which ingredients of the recipe carry the weight. We find that a
parameter EMA, rather than the last iterate, is consistently the
stronger deployment policy, and that top-advantage filtering,
training on only the highest-absolute-advantage fraction of collected samples,
halves the wall-clock cost of a training iteration.
\item \textbf{A JAX-native \generals{} environment}\footnote{The
environment is released as open source; available
at \url{https://github.com/strakam/generals-bots}.} that reaches tens of
millions of frames per second on a single GPU and supports all three
game modes through a shared action and observation interface, a
testbed for research in cooperative and general-sum multi-agent
reinforcement learning (all learning results in this paper are for
1v1).
\end{itemize}

\section{Related Work}
\label{sec:related}

\subsection{Policy-Gradient Methods in Zero-Sum Games}
Rudolph et al.~\cite{rudolph2025reevaluating} hypothesize that, with
proper tuning, generic policy-gradient methods are competitive with
or superior to specialized game-theoretic alternatives (CFR~\cite{cfr},
fictitious play~\cite{brown1951fictitious},
double-oracle~\cite{mcmahan2003double}, PSRO~\cite{psro}) in
imperfect-information games, and support this with a large
exploitability study on standard benchmarks. Sokota et
al.~\cite{sokota2025superhuman} and Perolat et al.~\cite{stratego}
apply the same idea at scale, using two regularized policy
gradients, magnetic mirror descent (MMD)~\cite{sokota2023unified}
and R-NaD respectively, to reach top-level play in Stratego. We add a case
study showing that policy-gradient self-play also scales to a large real-time
strategy game.

\subsection{Prior \generals{} Work}
Bhatia et al.~\cite{generally_genius} highlighted \generals{} as a
promising testbed for AI research and built a data-collection and
bot-integration framework, and used it to analyse the rule-based
agent \texttt{Flobot}. Xu et al.~\cite{levine} likewise framed
\generals{} as a cost-effective analogue of Dota~2 and StarCraft~II,
and proposed a hierarchical self-play agent (\texttt{HASP}) that
reaches a reported 77\% win rate against \texttt{Flobot}. Neither
released an RL-friendly environment supporting the vectorized parallel rollouts that large-scale training depends on. Both
agents are also far below the community-developed
\texttt{Human.exe}, the strongest \emph{non-learning} agent, which
wins close to 100\% of games against either. \texttt{Human.exe}
combines a wide range of graph algorithms and dynamic programming,
and was until recently the top bot on the leaderboard.
Straka and Schmid~\cite{straka2024generals} introduced a gym-like,
NumPy-based \generals{} environment and trained a PPO agent that
reached a top-25 placement on the 1v1 leaderboard, above
\texttt{Human.exe}, relying on behavior cloning from expert replays,
potential-based reward shaping, and population-based self-play.

\section{Preliminaries}
\label{sec:preliminaries}

We focus on 1v1 \generals{}, modeled as a two-player partially
observable stochastic game (POSG)~\cite{hansen2004posg}; the
formalism extends naturally to the 2v2 and free-for-all modes our
environment also supports (Section~\ref{sec:game-variants}). At each
turn $t$, the global state is $s_t$; each player $i \in \{0, 1\}$
observes $o_t^{(i)}$ rather than the full state and selects an action
$a_t^{(i)} \sim \pi_\theta(\cdot \mid h_t^{(i)})$ from a shared policy
used in self-play, where $h_t^{(i)} = \phi\big(o_1^{(i)}, \dots,
o_t^{(i)}\big)$ augments the current observation with memory of
earlier ones (Sec.~\ref{app:observation} of the supplementary material\footnote{This
article has supplementary downloadable material available at
\url{http://ieeexplore.ieee.org}, provided by the authors. It contains
the observation layout, hyperparameters, simulator benchmarks and
validation, and ablation details.}). The joint action
$a_t = (a_t^{(0)}, a_t^{(1)})$ drives a deterministic transition
$s_{t+1} = f(s_t, a_t)$; randomness enters only through the
initial-state distribution, which samples the map layout and
the two generals' positions. Imperfect information arises both from
the fog of war and from the simultaneity of moves: each player
chooses without observing the other's action. The reward is sparse,
terminal, and zero-sum: at the end of the game the winner receives
$r = +1$ and the loser $r = -1$.

The agent is trained by self-play to maximize the expected terminal
outcome $\mathbb{E}_{\tau \sim \pi_\theta}[r]$ with Proximal Policy
Optimization (PPO)~\cite{schulman2017ppo}, an on-policy
policy-gradient algorithm; we refer the reader unfamiliar with
reinforcement learning to Sutton and
Barto~\cite{sutton2018reinforcement} for a thorough introduction. At
each iteration we collect a batch of self-play trajectories under the
current policy $\pi_{\theta_k}$, estimate the advantage $\hat{A}_t$
of each action, and update $\theta$ by ascending the clipped
surrogate
\begin{equation}
\label{eq:ppo}
J(\theta) = \mathbb{E}_t\!\left[
\min\!\big(\rho_t(\theta)\,\hat{A}_t,\;
\mathrm{clip}(\rho_t(\theta), 1-\epsilon, 1+\epsilon)\,\hat{A}_t\big)
\right],
\end{equation}
where $\rho_t(\theta) = \pi_\theta(a_t \mid h_t) / \pi_{\theta_k}(a_t
\mid h_t)$ is the probability ratio between the updated policy and
the policy $\pi_{\theta_k}$ that collected the data. Clipping
$\rho_t(\theta)$ to $[1-\epsilon, 1+\epsilon]$ caps how far a single
update can move the policy from $\pi_{\theta_k}$, which keeps
on-policy updates stable. Because self-play makes the learner its own
opponent, the distribution it trains against shifts as $\theta$
changes, so we add an entropy bonus to keep the policy from
collapsing prematurely onto narrow strategies. We give the full
objective, including this regularizer and the value-function loss, in
Section~\ref{sec:agent-objective}.

\section{Game Description}
\label{sec:game}

\generals{} is a multiplayer real-time strategy game on a grid with
partial observability. Each player commands an army that grows over
time, expands across the grid, and tries to capture the opponent's
\emph{general} while defending their own. Resource allocation,
opponent modeling, and deception under fog of war together support a
wide range of strategies and emergent behaviors. The same rule set
covers 1v1, 2v2, and free-for-all formats
(Fig.~\ref{fig:game_modes}), exposed through our JAX-native
environment via a shared gym-like interface.

The rest of this section walks through the rules in detail: how
maps are generated, what each player sees, how armies grow, and how
movement and combat resolve. We then describe the team and
free-for-all variants, sketch the strategic depth the rules give
rise to, and close with the JAX environment we release alongside
the agent.

\begin{figure*}[t]
    \centering
    \includegraphics[width=0.95\textwidth]{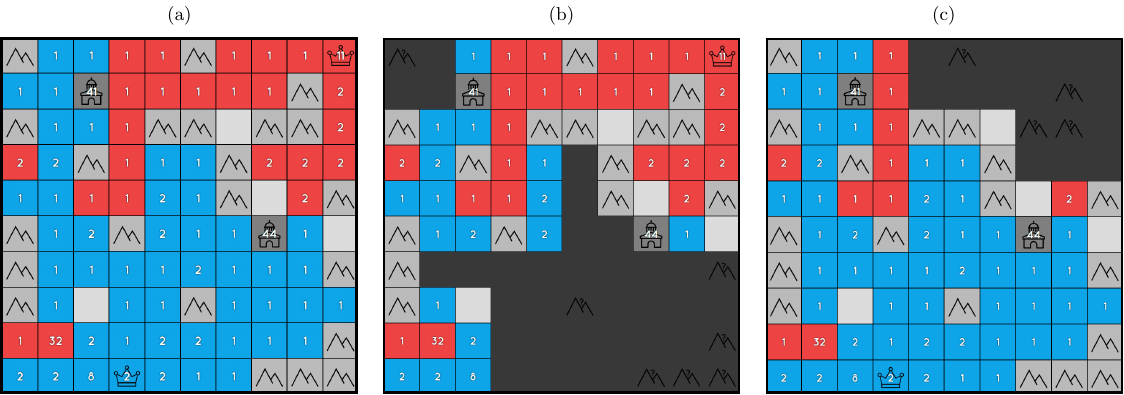}
    \caption{Three views of the same game state:
    (a) perfect-information view; (b) the red player's view; (c) the
    blue player's view. Crowns mark bases; numbers are unit counts.}
    \label{fig:three_views}
\end{figure*}

\subsection{Grid Generation}
\label{sec:game-grid}
In the official \generals{} game, matches are played on an
$H \times W$ grid (up to $23 \times 23$) populated with four cell
types: \emph{plain} (traversable), \emph{mountain} (impassable),
\emph{general} (a player's base), and \emph{castle}
(army-generating cell). The map generator fills a random $20\%$ of
cells with mountains and places $9$--$11$ neutral castles, each
pre-loaded with $40$--$50$ neutral units. The two generals are placed
at least $17$ steps apart, and each player is guaranteed at least one
neutral castle within radius $6$ of their general.

\subsection{Ownership and Partial Observability}
\label{sec:game-fog}
\begin{figure}[t]
    \centering
    \setlength{\fboxsep}{0pt}\setlength{\fboxrule}{1pt}%
    \fbox{\includegraphics[width=0.6\linewidth]{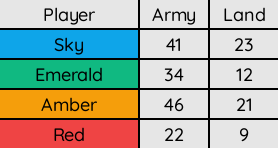}}
    \caption{The in-game scoreboard, exposing each player's total
    land (owned-cell count) and total army for the current turn.}
    \label{fig:scoreboard}
\end{figure}

Each cell is either neutral or owned by a single player. A fog-of-war
mechanic restricts each player's view to the cells they own together
with the Moore (8-cell) neighborhood around them; all other cells are
hidden (Fig.~\ref{fig:three_views}). In parallel, a public global scoreboard
(Fig.~\ref{fig:scoreboard}) exposes each player's total owned-cell
count and aggregate army count. By watching how these totals
evolve, both players can infer what the opponent is doing despite
the fog of war, for instance whether they are expanding or
consolidating, and bound the maximum army the opponent could have
stockpiled at their base.

\subsection{Army Growth}
\label{sec:game-growth}
Players grow their armies through two production mechanisms acting
on different timescales. Generals and owned castles produce one
unit every other turn. In addition, every $50$ turns the entire
territory ticks: each plain cell the player owns gains one unit
simultaneously. Capturing a neutral castle requires paying down its
$40$--$50$ starting units through combat; once captured, it produces
like any owned castle.

\subsection{Movement and Combat}
\label{sec:game-combat}
Gameplay advances in turns; in the browser version each turn lasts
$0.5$ seconds, and all players' moves are resolved
simultaneously. On each turn, a player selects one of their owned
cells and a cardinal direction, dispatching either all-but-one unit
or exactly half of the cell's units to the neighboring cell.
Players may also pass, leaving the board unchanged.

Combat is resolved by subtraction. If the destination is empty, the
dispatched units simply move there. If it is held by an enemy army,
the two stacks cancel one-for-one and whichever side is larger
survives with the remaining difference, taking ownership of the cell
(ties leave the cell with its previous owner). The win condition is
to capture the opponent's general: move an army onto the general's
cell and win the resulting combat.

\subsection{Team and Free-for-All Variants}
\label{sec:game-variants}
The same mechanics scale to multi-player formats with two natural
modifications.
In free-for-all with $N$ players, capturing another player's general
transfers all of their cells to the captor along with half of their
total army, and the captured general is converted into a regular
castle. The last general standing wins.
In team play (2v2), moving onto an ally-owned cell transfers the
cell's ownership to the mover while pooling the ally's units there
into the moving stack, rather than subtracting them as enemy units
would. Allies share the win/loss condition: the team loses when both
of its generals have been captured.

\subsection{Strategic Complexity}
\label{sec:game-strategy}
Several interacting mechanics give \generals{} its strategic depth.
\emph{Fog of war} injects persistent uncertainty: a player must
infer the opponent's general location, army composition, and intent
from local observations and the global scoreboard. \emph{Deception}
emerges naturally: feints toward one flank can mask a real attack
elsewhere, and a player can route an army around the opponent
through the fog and strike their general from behind once it is
left undefended.
\emph{Tempo} is central: capturing territory just before the
$50$-turn reinforcement tick yields a disproportionate payoff, while
over-expanding leaves the army too thin to repel a focused attack.
\emph{Snowballing} compounds early advantages, since each captured
cell contributes to future production, so small early mistakes can
cascade into game-deciding asymmetries. Capturing a castle is the
clearest instance of a longer-horizon investment: it yields a
permanent boost to production, but the army spent on the capture
leaves the player materially weaker for a stretch of turns, so the
agent must time the investment to a window when the opponent cannot
exploit the gap.

\subsection{JAX Environment}
\label{sec:game-env}
Prior research on \generals{} has relied either on a non-vectorizable
bot-integration framework~\cite{generally_genius} or on a NumPy-based
vectorized environment coupled to a PyTorch training
stack~\cite{straka2024generals}, the latter reporting around
$3{,}500$ frames per second with 12 parallel environments on a
12-core CPU. Throughout this paper, one \emph{frame} is one
environment advanced by one game turn; a batched call that advances
$N$ environments by one turn therefore counts as $N$ frames, and in
self-play each frame yields one training sample per player. We
reimplement the environment, the rollouts, and the training loop in
JAX, so that
the entire pipeline can be \texttt{jit}-compiled end-to-end and
dispatched to a GPU as a single graph. Observations, rewards,
fog-of-war masking, army production, combat resolution, and the
spawn-distance curriculum are all expressed as pure functions over
a flat \texttt{PyTree} of game state.

\begin{figure}[t]
    \centering
    \includegraphics[width=\linewidth]{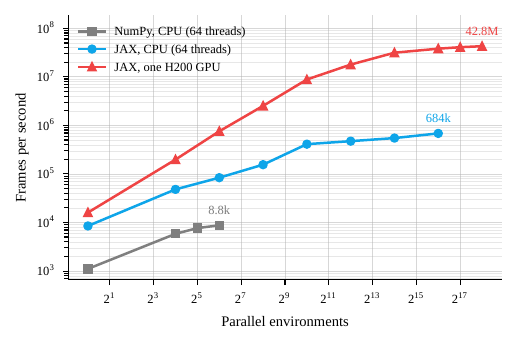}
    \caption{Simulator throughput against the number of parallel
    environments (both axes logarithmic). One frame is one environment
    advanced by one turn. The prior NumPy simulator~\cite{straka2024generals}
    and our JAX simulator are measured on one CPU with 64 threads
    (median of three runs); the JAX simulator is also measured on one
    H200 GPU. Peak values are annotated.}
    \label{fig:sim_throughput}
\end{figure}

\paragraph{Throughput}
Fig.~\ref{fig:sim_throughput} plots frames per second against the
number of parallel environments for three configurations: the prior
NumPy simulator~\cite{straka2024generals} and ours on the same
64-thread CPU node, and ours on a single NVIDIA H200 GPU (the NumPy
simulator cannot run on a GPU). Both
simulators receive uniformly random actions; the prior simulator runs
one process per thread, and ours is swept until throughput saturates.
On identical hardware the JAX implementation is about $78\times$ faster
at saturation ($684$k versus $8.8$k frames per second); on one H200
it reaches $42.8$M frames per second at $2^{18}$ parallel
environments, roughly $4{,}900\times$ the CPU peak of the prior simulator. The two-player game is the fastest configuration; 2v2 and
four-player free-for-all run at $18.6$M frames per second on the same
GPU. Hardware, software versions, and the complete measurements are
given in Sec.~\ref{app:sim-throughput} of the supplementary material; the benchmark
scripts are released with the environment.

\paragraph{Correctness}
The simulator is validated in three ways, detailed in Sec.~\ref{app:sim-tests} of the
supplementary material: a suite of $145$ unit and property tests
covering every rule of the game; replay of $10{,}000$ ranked games,
drawn at random from the deployment window until September 20, 2026,
all of which replay perfectly against the version of the game live on
that date, meaning that the simulated board agrees with the game's own
engine on every cell of every turn; and the agent's record over
more than a thousand rated games on the live server, played through the same
observation pipeline.

\paragraph{Interface}
The environment exposes the same observation and action interface for
the 1v1, 2v2, and free-for-all formats. The action space is $H \times
W \times 9$ per player in every mode (a per-cell choice of pass, or
one of two move types, send all but one unit or send half, in each of the
four cardinal directions), and the observation tensor encodes the
agent's view of the board (Sec.~\ref{app:observation} of the supplementary material).
All learning results in this paper are for 1v1; learning in the team
and free-for-all modes is future work.

\section{Agent}
\label{sec:agent}

This work introduces an AI for \generals{} trained from random initialization by
self-play reinforcement learning, without human demonstrations and
without hand-designed reward shaping. Both players are controlled
by the same network with the same parameters, and every self-play
game contributes both players' transitions to the update; the
opponent is the current iterate. Games that reach $2{,}048$ turns are
truncated with reward $0$ to both players and counted as draws in
evaluation. The remainder of this section describes the components
of the agent.

\subsection{Network Architecture}
\label{sec:agent-arch}
\begin{figure*}[t]
    \centering
    \includegraphics[width=\textwidth]{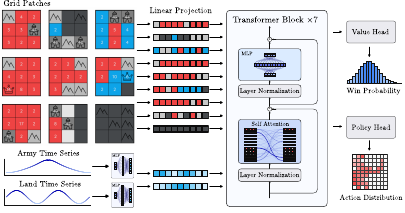}
    \caption{The environment observation is spatially sliced into
    $3\times 3$ patches and linearly projected into tokens. Alongside
    these, two time series of land and army counts are each projected
    by an MLP into the token space. All tokens pass through the
    transformer and are then consumed by a value head that predicts a
    win-probability distribution and a policy head that produces a
    distribution over all possible actions.}
    \label{fig:pipeline}
\end{figure*}

Fig.~\ref{fig:pipeline} shows the overall architecture. The policy is
parameterized by a transformer torso. The environment produces a
feature tensor $o \in \mathbb{R}^{C \times H \times W}$, where $H$
and $W$ are the grid's height and width and $C = 38$ channels encode
the current observation together with a memory augmentation (for
example, the location of the opponent's base once it has been
spotted). The spatial tensor is split into $3 \times 3$ patches,
and each patch is embedded into a single input token. Two additional
non-spatial tokens carry global temporal statistics rather than
per-cell features: a 512-step sliding window of the opponent's total
army count and a matching window of their total land count, each
projected by an MLP into a single token. These trajectories let the
agent infer what the opponent is doing behind the fog of war,
whether they are expanding or consolidating, from the way the
scoreboard totals evolve.

Value and policy heads sit on top of the torso; the policy head
produces an $H \times W \times 9$ distribution that, for each source
cell, assigns a probability to each of nine actions: pass, or one of
two move types (send all but one unit or send half) in each of the four
cardinal directions ($1 + 2 \times 4 = 9$). Transformer
hyperparameters are listed in Table~\ref{tab:arch};
Sec.~\ref{app:observation} of the supplementary material gives the observation-channel
layout.

\begin{table}[t]
    \centering
    \caption{Network hyperparameters.}
    \label{tab:arch}
    \small
    \renewcommand{\arraystretch}{1.3}
    \rowcolors{2}{white}{gray!8}
    \begin{tabular}{ll}
        \toprule
        \textbf{Parameter} & \textbf{Value} \\
        \midrule[\heavyrulewidth]
        Depth                 & $7$ \\
        Embedding dimension   & $448$ \\
        Feedforward dimension & $1344$ \\
        Attention heads       & $8$ \\
        Total parameters      & $15.35$M \\
        \bottomrule
    \end{tabular}
\end{table}

\subsection{Training Objective}
\label{sec:agent-objective}
Our training objective is a PPO loss, combining the on-policy
policy-gradient surrogate $J(\theta)$ of Section~\ref{sec:preliminaries}
with an entropy bonus and a value-function loss:
\begin{equation}
\label{eq:loss}
\mathcal{L}(\theta) = -J(\theta)
\;-\;\alpha\,H(\pi_\theta)
\;+\;\beta\,\mathcal{L}_{\mathrm{value}}(\theta),
\end{equation}
where $H(\pi_\theta)$ is the policy entropy. The entropy bonus
encourages exploration and keeps the policy from collapsing
prematurely onto narrow strategies. We also experimented with
regularizing toward several non-parametric heuristic policies in
place of the entropy bonus, but observed no improvement, so we kept
plain entropy for simplicity. Adding a further KL regularizer
toward the previous iterate
$\mathrm{KL}\!\left(\pi_\theta\,\Vert\,\pi_{\mathrm{old}}\right)$
would recover magnetic mirror descent~\cite{sokota2023unified}; we
omit it, which saves one forward pass through $\pi_{\mathrm{old}}$
per minibatch and in our experiments leaves training stable. For
$\mathcal{L}_{\mathrm{value}}$ we use the categorical HL-Gauss
value loss of Farebrother et al.~\cite{farebrother2024hlgauss}.

\subsection{Spawn-Distance Curriculum}
\label{sec:agent-curriculum}
In full-size \generals{}, the minimum BFS distance between the two
generals is constrained to be at least $17$ steps, and training
directly in this regime is not effective: a randomly initialized
policy rarely stumbles onto capturing the opponent's base, so the
win/loss signal is too sparse to bootstrap learning. We therefore run
a curriculum over spawn distance: the BFS distance between the two
generals is drawn from a window that starts at $2$--$6$ cells and
widens over five stages ($4$--$8$, $6$--$13$, $11$--$17$, and finally
$17$--$28$). A stage advances when the current greedy policy wins at
least $60\%$ of $512$ evaluation games against an opponent that
plays uniformly random moves.

\subsection{Exponential Moving Average}
\label{sec:agent-ema}
We accumulate the policy parameters over the course of training into
an EMA $\bar\theta_t = \tau\bar\theta_{t-1} + (1-\tau)\theta_t$, as
Sokota et al.~\cite{sokota2025superhuman} also do.
Training gradients act on $\theta$; at deployment we evaluate
$\bar\theta$. Our intuition is twofold: the average smooths out the
abrupt changes that a single training iteration makes to the policy,
and because it spans a longer stretch of training (with
$\tau = 0.999$, roughly the last thousand iterations), it retains
behaviour against opponents that the current iterate has since
stopped meeting in self-play, which should make the deployed policy
robust to a wider range of opponents. Section~\ref{sec:ablations-ema}
measures the difference between $\bar\theta$ and $\theta$.

\subsection{Top-Advantage Filtering}
\label{sec:agent-topadv}
Inspired by Sokota et al.~\cite{sokota2025superhuman}, we train on
only a fraction $k$ of the transitions of each rollout batch, those
with the largest advantage magnitude. After computing the advantages with generalized advantage
estimation~\cite{schulman2016gae} we normalize them over the whole
batch, rank the
transitions by $|\hat{A}_t|$, and keep the top fraction $k$. All three
terms of Eq.~(\ref{eq:loss}) are computed on the kept
transitions only, with
the minibatch size unchanged, so an iteration takes a fraction $k$ of
the optimizer steps of training on the whole batch. Because the
ranking is by magnitude, negative-advantage transitions are kept and
the filter does not push the policy toward the selected actions. The
final agent
uses $k = 25\%$; Section~\ref{sec:ablations-topadv} sweeps $k$.

\section{Evaluation}
\label{sec:eval}

We evaluate the agent on the public 1v1 ladder of \generals{}, where
the matchmaker assigns rated games against the community, including
the leading human players and the strongest prior bots. The site
rates players with OpenSkill~\cite{openskill}, a Bayesian rating
system in the TrueSkill~\cite{herbrich2006trueskill} family that models each player's skill as a
Gaussian with mean $\mu$ and uncertainty $\sigma$, and displays the
conservative rating $\mu - 3\sigma$ as stars: a lower bound on the
skill estimate that rises as a player's games accumulate and
$\sigma$ shrinks. We report four
records: the official weekly standings, ratings recomputed from every
ranked game played on the ladder in a two-month window, the agent's
own games in that window, and its head-to-head records against the
strongest humans and the prior bots.

\begin{figure}[t]
    \centering
    \includegraphics[width=\linewidth]{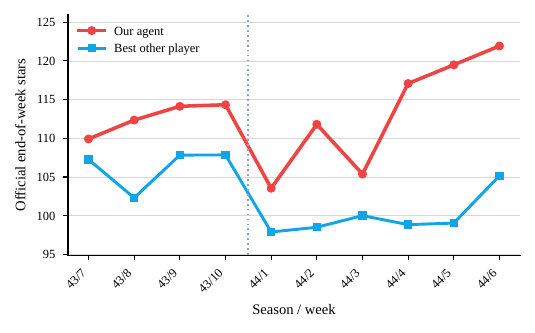}
    \caption{Official end-of-week standings for the ten weeks that
    overlap the agent's two-month deployment: our agent (red) is first
    in all ten, blue
    is the best other player. The dotted line separates the seasons.}
    \label{fig:leaderboard}
\end{figure}

\subsection{Prior Bots}
\label{sec:eval-baselines}
Two prior bots play on the public ladder: the heuristic agent
\texttt{Human.exe} and the learned agent \texttt{zero v3}.

\paragraph{\texttt{Human.exe}~\cite{humanexe}}
A rule-based agent that tries about thirty plans in a fixed
priority order each turn and plays the first that applies. Its
planners use prize-collecting Steiner trees for gathering army,
min-cost flow over regions of contiguous tiles for expansion, and
knapsack optimization to combine candidate paths within a turn
budget; engagements are resolved by an intercept planner over
chokepoints. A belief state tracks armies that entered the fog along
several candidate routes and infers hidden tile counts, army totals,
and the general's location from scoreboard changes and the points
where armies reappear.

\paragraph{\texttt{zero v3}~\cite{straka2024generals}}
A convolutional U-Net trained by behavior cloning, PPO self-play
with potential-based reward shaping, and population-based self-play
with three models. It reaches the public ladder's top $25$ and wins
$54.8\%$ of $529$ games against \texttt{Human.exe}.

\subsection{Protocol}
\label{sec:eval-protocol}
We deployed the EMA parameters of the final run on the ladder for a
two-month window under two accounts, \texttt{Average Joe}\footnote{Replays:
\url{https://generals.io/profiles/Average\%20Joe}} and
\texttt{L\_7d\_gae90\_30k\_ema}\footnote{Replays:
\url{https://generals.io/profiles/L\_7d\_gae90\_30k\_ema}}, with the
same checkpoint throughout (two accounts to gather games at double
rate). A laptop CPU ran the public client at the
game's pace of one action per $0.5$-second tick (Sec.~\ref{sec:supp-ladder}).

The official standings are the site's end-of-week tables for the ten
weeks of the deployment. Because the site's displayed rating depends
on a hidden uncertainty parameter ($\sigma$) that is not part of the
public record, we collected every ranked game played on the ladder in
the window and recomputed Elo and OpenSkill ratings for all players
from a common start (Sec.~\ref{sec:supp-reconstruction}). The agent's own record in the window
holds $1{,}147$ rated games; games between our two accounts and
unrated custom games are excluded, except for a custom series against
\texttt{zero v3}. Every game is publicly recorded, and all replays are
released with the code (Sec.~\ref{sec:supp-ladder}).

\subsection{Results}
\label{sec:eval-h2h}
\paragraph{Official standings}
The agent is first in every one of the ten weekly standings that
overlap its deployment (Fig.~\ref{fig:leaderboard}), by $2.7$ to
$20.4$ stars over the best other player; its stars peak at $121.9$, and the highest value any
other player reaches in these snapshots is $107.9$.
The snapshots are one longitudinal record, not ten independent tests.

\paragraph{Recomputed ratings}
Over the $173{,}776$ ranked games of the window among $14{,}616$
players (alternate accounts merged where the owner confirmed them),
we recompute chronological Elo and OpenSkill ratings from a common
start (Sec.~\ref{sec:supp-reconstruction}). The agent is first under every rating and setting;
Table~\ref{tab:ratings} gives its lead and the chance that another
player is rated at least as high. $K$ sets how many recent games a rating reflects,
about $400/K$; $K = 10$ is FIDE's value for established players
rated $2{,}400$ and above~\cite{fide2024rating,elo1978rating}. Across the whole window
the agent is the strongest player on the ladder under every rating
system and setting, with over $95\%$ confidence in five of the six.

\paragraph{Own games}
In the window the agent wins $946$ of $1{,}147$ games ($82.5 \pm 2.2\%$),
including $69.6\%$ against humans rated $100$ and above
(Table~\ref{tab:h2h}); against every opponent-rating band its $95\%$
lower bound exceeds $50\%$.

\paragraph{Top players and prior bots}
Against the two highest-rated humans, over all their
accounts, we extend the record past the window to 2026-09-20. The agent
is $189$--$67$ against \texttt{Mithraaaa}, regarded by the community as
the best player of all time (one-sided binomial $p < 10^{-14}$), and
$32$--$18$ against \texttt{shimatetsu} ($p = 0.032$); the two share the
highest rating any human reached on the ladder in the window. Of these
games, $7$ and $8$ were played after the window with the same
checkpoint. Against the
prior bots it is $13$--$0$ against \texttt{Human.exe} in rated games
and $22$--$1$ against \texttt{zero v3} in custom games; the one
loss was a connection failure on our side (Sec.~\ref{sec:supp-ladder}).

Taken together, the four records above---first in every official
standing, first under every recomputed rating, ahead of every rating
interval of the field, and significantly ahead of each of the two
strongest humans---are the sense in which we call the agent
superhuman.

\begin{table}[t]
    \centering
    \caption{Recomputed ratings over the window. Lead: margin over the
    strongest other player. $P_{\rm all}$, $P_{100}$: chance that another
    player is rated at least as high, among all players and among the
    $571$ with at least $100$ games. For Elo we estimate it by resampling
    whole days of the ladder; OpenSkill gives it directly (Sec.~\ref{sec:supp-reconstruction}).
    Bold: below $5\%$.}
    \label{tab:ratings}
    \footnotesize
    \setlength{\tabcolsep}{6pt}
    \begin{tabular}{lrrr}
        \toprule
        \textbf{Rating} & \textbf{Lead} & $\boldsymbol{P_{\rm all}}$ & $\boldsymbol{P_{100}}$ \\
        \midrule
        Elo, $K = 10$ & $154$ & $\mathbf{0.3\%}$ & $\mathbf{0.2\%}$ \\
        Elo, $K = 20$ & $137$ & $6\%$ & $\mathbf{3\%}$ \\
        Elo, $K = 40$ & $156$ & $25\%$ & $11\%$ \\
        OpenSkill, $\tau = 0$ & $8.8\,\mu$ & -- & $\mathbf{2.3\%}$ \\
        OpenSkill, $\tau = 1/12$ & $10.7\,\mu$ & -- & $\mathbf{1.2\%}$ \\
        OpenSkill, $\tau = 1/6$ & $13.2\,\mu$ & -- & $\mathbf{0.7\%}$ \\
        \bottomrule
    \end{tabular}
\end{table}

\begin{table}[t]
    \centering
    \caption{The agent's rated games in the window by opponent rating
    (human opponents), its games against the two strongest humans over
    all their accounts, and its games against the prior bots. LB:
    one-sided $95\%$ Clopper--Pearson lower bound on the win rate.}
    \label{tab:h2h}
    \footnotesize
    \setlength{\tabcolsep}{4pt}
    \renewcommand{\arraystretch}{1.15}
    \begin{tabular}{lrrrrc}
        \toprule
        \textbf{Opponents} & \textbf{Games} & \textbf{W} & \textbf{L} & \textbf{Win} & \textbf{95\% LB} \\
        \midrule[\heavyrulewidth]
        Rated $\geq 100$ & $158$ & $110$ & $48$ & $69.6\%$ & $63$ \\
        Rated $90$--$99$ & $330$ & $263$ & $67$ & $79.7\%$ & $75$ \\
        Rated $80$--$89$ & $235$ & $198$ & $37$ & $84.3\%$ & $79$ \\
        Rated $< 80$     & $346$ & $306$ & $40$ & $88.4\%$ & $85$ \\
        All humans       & $1{,}069$ & $877$ & $192$ & $82.0\%$ & $79$ \\
        \midrule
        \texttt{shimatetsu}$^{a,d}$ & $50$ & $32$ & $18$ & $64.0\%$ & $51$ \\
        \texttt{Mithraaaa}$^{b,d}$  & $256$ & $189$ & $67$ & $73.8\%$ & $68$ \\
        \midrule
        \texttt{Human.exe}        & $13$ & $13$ & $0$ & $100\%$ & $79$ \\
        \texttt{zero v3} (custom)$^{c}$ & $23$ & $22$ & $1$ & $95.7\%$ & $80$ \\
        \bottomrule
        \multicolumn{6}{@{}l@{}}{\scriptsize $^{a}$\texttt{shimatetsu}, \texttt{Reisen}. $^{b}$\texttt{Mithraaaa}, \texttt{Sinyth}, \texttt{Perihelion}.} \\
        \multicolumn{6}{@{}l@{}}{\scriptsize $^{c}$The loss was a connection failure on our side.} \\
        \multicolumn{6}{@{}l@{}}{\scriptsize $^{d}$To 2026-09-20; $8$ and $7$ games after the window (Sec.~\ref{sec:supp-ladder}).}
    \end{tabular}
\end{table}

\addtocounter{figure}{2}%
\begin{figure*}[t]
    \centering
    \includegraphics[width=0.95\textwidth]{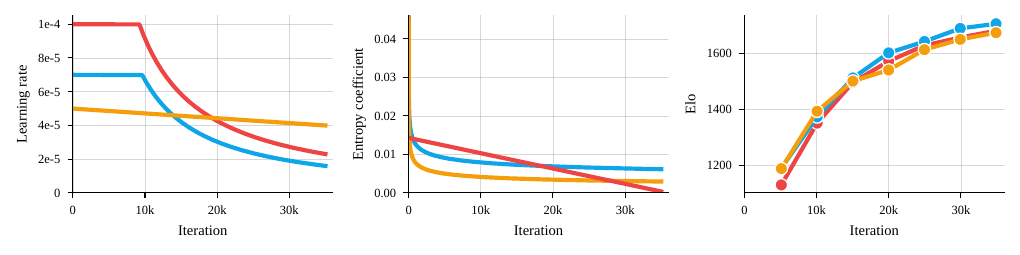}
    \caption{Relationship between different schedules and Elo.
    \textbf{Left:} learning-rate schedules---power-law decays (blue,
    red) and a linear decay (yellow). \textbf{Middle:}
    entropy-coefficient schedules---power-law decays (blue, yellow)
    and a linear decay (red). \textbf{Right:} local Elo of the three
    runs, from a round-robin among their checkpoints.}
    \label{fig:schedules_elo}
\end{figure*}
\addtocounter{figure}{-3}%
\begin{figure}[t]
    \centering
    \includegraphics[width=\linewidth]{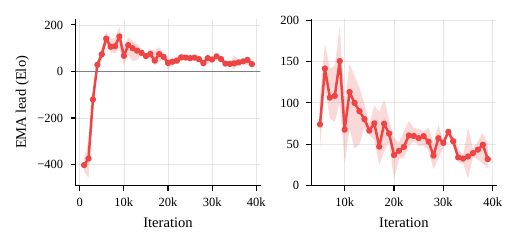}
    \caption{Elo lead of the EMA parameters over the last iterate of the
    same run at the same iteration, from $500$ direct games per point:
    mean over the three large-scale runs (line) and the range across
    them (band). \textbf{Left:} every $1{,}000$ iterations from $1{,}000$
    to $39{,}000$. \textbf{Right:} the same from $5{,}000$ on.}
    \label{fig:ema}
\end{figure}

\begin{figure}[t]
    \centering
    \includegraphics[width=\linewidth]{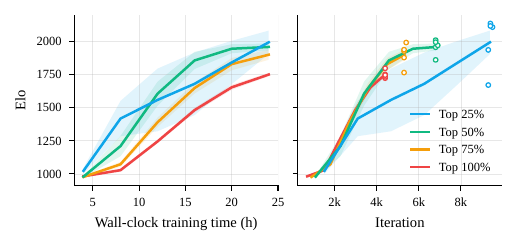}
    \caption{Top-advantage filtering ablation: tournament Elo of the
    checkpoints taken every four hours of training, against
    wall-clock time (left) and against iterations (right). Mean $\pm$
    standard error over five seeds per setting; small markers mark each
    run's final checkpoint. Filtering
    is ahead at equal wall-clock time and level at equal frames.}
    \label{fig:topadv_wallclock}
\end{figure}
\addtocounter{figure}{1}%

\section{Ablations}
\label{sec:ablations}

We isolate several design choices from Section~\ref{sec:agent}: the
deployment policy (Section~\ref{sec:ablations-ema}), the top-advantage
filtering fraction (Section~\ref{sec:ablations-topadv}), and the reward
signal (Section~\ref{sec:ablations-reward}). We
additionally report robustness to learning-rate and
entropy-coefficient schedules (Section~\ref{sec:ablations-schedules})
and an analysis of one attention head
(Section~\ref{sec:ablations-attention}). All local Elo values in this
section come from round-robin tournaments: the checkpoints compared in
a figure play every other checkpoint in that figure in both seats, and
a single Elo fit per figure places them on one scale (Sec.~\ref{app:ablation-details} of
the supplementary material).

\subsection{Exponential Moving Average}
\label{sec:ablations-ema}
Across larger-scale experiments we ran during development, the
EMA of the policy parameters was stronger at inference than the
corresponding last iterate (Fig.~\ref{fig:ema}). On each of our
three large-scale runs, the EMA plays the last iterate of the same
iteration in $500$ games, every $1{,}000$ iterations up to $39{,}000$
(Sec.~\ref{app:ablation-details}). For the first few
thousand iterations the EMA is behind. With decay $0.999$ it averages
over roughly the last $1{,}000$ iterations, so early in training it
still carries much of the weak parameters of the first iterations,
including the initial ones and those learned on the first curriculum
stages; the weight of the initialization, $0.999^{t}$, is still a third
at iteration $1{,}000$. From $4{,}000$ iterations on, the EMA is ahead
at every checkpoint of every run, by $65$ Elo on average. The lead
shrinks as training goes on but remains significant: at $39{,}000$
iterations the EMA is still $32$ Elo ahead (one-sided sign test over
its $1{,}500$ games, $p < 10^{-3}$). We find it notable that prior large-scale
self-play work on StarCraft~II~\cite{alphastar}, Dota~2~\cite{dota2},
or Stratego~\cite{stratego} does not discuss parameter EMA as a way to
improve the deployed policy, although it costs almost nothing to add;
the one exception we are aware of is Sokota et al.~\cite{sokota2025superhuman}. Understanding when and why
parameter EMA helps in self-play RL, whether it is a generic
stabilizer or specific to certain regimes, is a natural question for
future work.

\subsection{Top-Advantage Filtering}
\label{sec:ablations-topadv}
\begin{figure*}[t]
    \centering
    \includegraphics[width=0.95\textwidth]{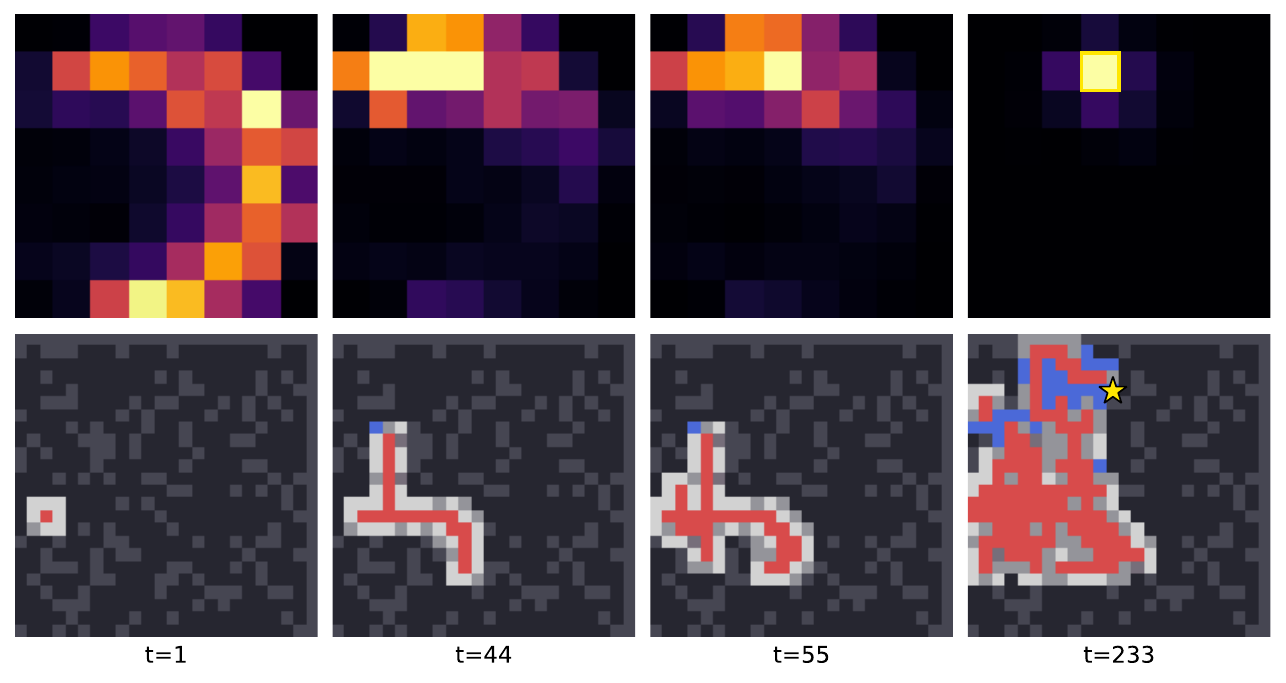}
    \caption{Value-token attention from head~4 of layer~5 (zero-indexed)
    at four snapshots from a single self-play game ($t = 1, 44, 55, 233$),
    shown alongside the player-0 board view. Each attention cell
    represents a $3 \times 3$ board patch. Attention initially forms
    a broad spatial band, narrows after first contact, and concentrates
    on the opponent general's patch after its location is revealed (star).}
    \label{fig:attention_head}
\end{figure*}

We ablate the top-advantage filter of Section~\ref{sec:agent-topadv}
by sweeping the retained fraction $k$ ($25$, $50$, $75$, $100\%$)
with five seeds per setting, $24$ hours on one H200 each. From every
run we take the checkpoints at $4, 8, \ldots, 20$ hours and the final
one, and rate all $120$ in one round-robin (Sec.~\ref{app:ablation-details}).
Fig.~\ref{fig:topadv_wallclock} plots the mean and standard error
across seeds against wall-clock time (left) and against iterations
(right); an iteration consumes the same number of environment frames
in every setting, so the right panel compares equal frame budgets.

Filtering helps significantly at equal wall-clock time: after $24$
hours the filtered runs end $150$ to $250$ Elo above the unfiltered
ones, and $13$ of the $15$ filtered runs end above every unfiltered
run (one-sided rank-sum test over seeds, $p = 0.002$). Per environment
frame, the settings learn about equally fast
(Fig.~\ref{fig:topadv_wallclock}, right). The gain comes from cheaper
iterations. Many frames come from positions whose outcome is already
decided, where advantages are near zero; the filter drops them, so an
iteration takes a fraction $k$ of the optimizer steps and the time
saved goes into fresh rollouts of the improved policy. In $24$ hours,
$k = 0.25$ completes twice as many iterations as $k = 1$. We do not
claim that $k = 0.25$ is the best setting; on its own it is not
significant ($p = 0.08$, as one seed ends low). We chose it for the
final agent because it was the strongest setting in our development
runs and three of its five seeds here end above every other run.
Sec.~\ref{app:topadv_sample} gives every setting and seed.

\subsection{Reward Shaping}
\label{sec:ablations-reward}
The previous state of the art, \texttt{zero v3}~\cite{straka2024generals},
was trained with potential-based reward shaping~\cite{ng1999pbrs}. Its
authors motivate shaping by the long episodes and their compute budget,
and report that without it self-play converged to an overly aggressive,
exploitable strategy; the shaped agent won $71.9\%$ of games against the
unshaped one. Our agent is trained on the sparse win/loss reward alone
and reaches superhuman play, although
its self-play games are long, about $730$ turns on average. In
our setting the reward needs no hand-designed game knowledge, even at
this episode length.

In our experiments, shaping does speed up early training. In a
controlled ablation with four seeds per reward on one GPU each
(Sec.~\ref{app:ablation-details}, Fig.~\ref{fig:reward}), the shaped runs lead by $150$ Elo on average
after $24$ hours of training. From there the lead decreases with
further training, to $90$ Elo after $48$ hours, when the best sparse
run is rated above two of the four shaped runs. The
final agent trains on roughly an order of magnitude more frames than
this ablation, and we hypothesize that with more frames, from longer
training on more GPUs, the advantage of shaping narrows further.

\subsection{Schedule Sensitivity}
\label{sec:ablations-schedules}
Rudolph et al.~\cite{rudolph2025reevaluating} report that carefully
tuned policy-gradient methods are comparable to or better than CFR-,
fictitious-play-, and PSRO-based methods across a range of imperfect-information
games. Building on that, Sokota et al.~\cite{sokota2025superhuman}
claim that the specific annealing schedules of the learning rate
and the entropy coefficient were critical for smooth training
dynamics. Over the three schedule pairs we tested, one run each, combinations
of power-law and linear decays for both the learning rate and the
entropy coefficient reach comparable local Elo
(Fig.~\ref{fig:schedules_elo}). We suspect that the
\emph{shape} of a schedule matters less than its \emph{range}: in our
experiments, annealing the learning rate from $10^{-4}$ to about
$2 \times 10^{-5}$ was safe. This evidence covers the three schedules
shown; a broader sensitivity analysis remains for future work.

\begin{table}[t]
    \centering
    \caption{Attention of head~4, layer~5 over $2{,}048$ self-play
    games, by game phase: the fraction of turns on which the attention
    peak lies on the opponent general's patch, the fraction on which
    it lies within one patch of it, and the mean distance in patches
    (Chebyshev) from the peak to the general's patch.}
    \label{tab:attention}
    \footnotesize
    \renewcommand{\arraystretch}{1.25}
    \rowcolors{2}{white}{gray!8}
    \begin{tabular}{@{}lrrr@{}}
        \toprule
        \textbf{Phase} & \textbf{On patch} & \textbf{Within one} & \textbf{Dist.\ (patches)} \\
        \midrule[\heavyrulewidth]
        Before contact          & $3\%$  & $23\%$ & $3.1$ \\
        Contact, general unseen & $32\%$ & $79\%$ & $1.1$ \\
        General revealed        & $94\%$ & $99\%$ & $0.1$ \\
        \bottomrule
    \end{tabular}
\end{table}

\subsection{Attention Head Analysis}
\label{sec:ablations-attention}
We were curious whether anything interpretable could be extracted
from a trained policy, in particular whether individual attention
heads attach to game-state features that match a human's notion of
what matters. The first-order answer is no: representations are
largely distributed, and most heads lack clean interpretability. We also noticed that when the game state changes
qualitatively (e.g., first contact with an enemy unit, first castle
capture), the apparent function of many heads shifts.

However, one head shows a spatial pattern consistent with a learned belief
over the opponent's general location: attention from the value token
to spatial patches in head~4 of layer~5 (Fig.~\ref{fig:attention_head}).
Each patch represents a $3 \times 3$ block of board cells.
The figure illustrates how this pattern develops during one self-play
game. At $t=1$, attention forms a broad band around the agent's
starting position, consistent with the map-generation constraint that
generals spawn at least $17$ BFS steps apart. At $t=44$, after first
contact with the enemy, attention concentrates into roughly a third
of the map, with three prominent candidate regions. At $t=55$,
without further sightings of the opponent, attention becomes more
concentrated around one candidate patch. One possible interpretation
is that the absence of an attack provides evidence against the closer
candidates: an opponent starting there could already have reached
the agent. At $t=233$, once the opponent's general is directly
observed, attention concentrates on the patch containing its location.

We quantify this over $2{,}048$ self-play games ($1.5$ million
turns): at every turn we compare the peak of this head's attention
over the $8 \times 8$ patches with the opponent's true general, split
by whether the player has met the enemy and whether the general has
been revealed (Table~\ref{tab:attention}). Once contact is made and
while the general is unseen, the peak is on the correct patch on
$32\%$ of the turns and within one patch on $79\%$ (a uniformly
random guess among the $64$ patches would give $1.6\%$ and $12\%$),
improving with time since contact. Ranked by the average probability their
attention places on the general's patch, this head is first of the
$56$ heads in the network. These localization results support
interpreting the head as representing plausible general locations;
they do not establish its causal role in the policy. Scripts, data,
and animations of $100$ games are released with the code.

\section{Conclusion}
\label{sec:conclusion}

In this work we have shown that, given enough compute, a self-play
policy-gradient learner reaches superhuman play in \generals{} without
human demonstrations, reward shaping or population-based self-play:
over a two-month deployment our
agent held the top of the public 1v1 ladder, led every recomputed
rating, and beat the strongest humans and prior bots. Two inexpensive
additions made a measurable difference: an EMA of the parameters gives
the stronger policy at deployment, and top-advantage filtering doubles
the iterations completed in a fixed training time. We release the
training code and a JAX-native simulator that also supports 2v2 and
free-for-all play; learning in these team and multi-player settings is
left for future work.

\bibliographystyle{IEEEtran}
\bibliography{main}

\clearpage
\section*{Supplementary Material}
\setcounter{section}{0}
\renewcommand{\thesection}{S-\Roman{section}}
\renewcommand{\theHsection}{S.\arabic{section}}
\setcounter{table}{0}
\renewcommand{\thetable}{S.\Roman{table}}
\renewcommand{\theHtable}{S.\arabic{table}}
\setcounter{figure}{0}
\renewcommand{\thefigure}{S.\arabic{figure}}
\renewcommand{\theHfigure}{S.\arabic{figure}}

\section{Observation Space}
\label{app:observation}

The environment exposes a feature tensor with
$C = 24 + 2H_{\text{hist}}$ channels, where $H_{\text{hist}}$ is the
per-cell delta history length (default $7$, giving $C = 38$
channels). All channels are $(H, W)$ spatial maps; scalar quantities
are broadcast across the grid, and cells outside the playable area
are padded with mountain tokens. Channels $0$--$3$ and $9$--$13$ are
instantaneous (read from the current step); channels $4$--$8$ and
$20$--$21$ are persistent memory accumulated across the episode;
channels $16$--$19$ are scalar game-state statistics broadcast to
every cell; and channels $24$ onward stack per-cell army-count deltas
over the last $H_{\text{hist}}$ steps. On top of the spatial
channels, the transformer also consumes two non-spatial temporal
tokens encoding the opponent's army-total and land-count time series
over a $512$-step sliding window; these are prepended as summary
tokens alongside the value token.

\begin{table*}[t]
    \centering
    \caption{Observation channels. Indices assume $H_{\text{hist}}=7$.}
    \label{tab:observation}
    \rowcolors{2}{white}{gray!8}
    \small
    \begin{tabular}{@{}llp{0.7\textwidth}@{}}
        \toprule
        \textbf{Index} & \textbf{Name} & \textbf{Encoding} \\
        \midrule
        $0$ & \texttt{armies} & Raw army count on each visible cell ($0$ if fogged). \\
        $1$ & \texttt{own\_army} & Army count on cells owned by the agent; $0$ elsewhere. \\
        $2$ & \texttt{enemy\_army} & Army count on cells owned by the opponent; $0$ elsewhere. \\
        $3$ & \texttt{neutral\_army} & Army count on neutral cells (unclaimed castles/generals); $0$ elsewhere. \\
        $4$ & \texttt{seen} & Persistent visibility mask: $1$ for any cell the agent has ever seen (max-pooled $3\times 3$ owned-cell halo). \\
        $5$ & \texttt{enemy\_seen} & Persistent mask: $1$ for any cell where an enemy unit has ever been observed. \\
        $6$ & \texttt{generals} & Persistent mask of all general positions ever revealed (own and opponent). \\
        $7$ & \texttt{castles} & Persistent mask of all castle positions ever revealed. \\
        $8$ & \texttt{mountains} & Persistent mountain mask (visible mountains $\cup$ known padded border). \\
        $9$ & \texttt{neutral\_cells} & Current-frame mask of neutral-owned cells. \\
        $10$ & \texttt{owned\_cells} & Current-frame mask of agent-owned cells. \\
        $11$ & \texttt{opponent\_cells} & Current-frame mask of opponent-owned cells. \\
        $12$ & \texttt{fog\_cells} & Current-frame fog-of-war mask (cells not visible this step). \\
        $13$ & \texttt{structures\_in\_fog} & Mask of fog cells known to contain a structure (castle/mountain/general) or unseen padded border. \\
        $14$ & \texttt{timestep} & Absolute game step, broadcast as a constant map. \\
        $15$ & \texttt{timestep\_mod50} & $(\text{timestep} \bmod 50)/50$, broadcast: phase within the army-production cycle. \\
        $16$ & \texttt{own\_land\_count} & Scalar: number of cells the agent owns, broadcast. \\
        $17$ & \texttt{own\_army\_count} & Scalar: total agent army, broadcast. \\
        $18$ & \texttt{opp\_land\_count} & Scalar: number of cells the opponent owns, broadcast. \\
        $19$ & \texttt{opp\_army\_count} & Scalar: total opponent army (last observed), broadcast. \\
        $20$ & \texttt{last\_enemy\_army\_seen} & Per-cell memory: most recently observed enemy army count at that cell (sticky, not reset when the cell re-fogs). \\
        $21$ & \texttt{last\_enemy\_army\_age} & Per-cell log-decayed age $\log(1+\Delta t)/5$ where $\Delta t$ is the number of steps since the enemy was last seen there ($0$ when currently visible). \\
        $22$ & \texttt{coord\_x} & Normalized column coordinate in $[0,1]$, broadcast across rows. \\
        $23$ & \texttt{coord\_y} & Normalized row coordinate in $[0,1]$, broadcast across columns. \\
        $24$--$30$ & \texttt{own\_army\_delta} & Stack of $H_{\text{hist}}$ per-cell differences $\texttt{own\_army}(t-k)-\texttt{own\_army}(t-k-1)$. \\
        $31$--$37$ & \texttt{enemy\_army\_delta} & Stack of $H_{\text{hist}}$ per-cell differences of enemy army counts. \\
        \bottomrule
    \end{tabular}
\end{table*}

\paragraph{Memory update rules}
The persistent channels are maintained per player by an observation
state that is reset to zero when a game ends. Channel 4 (\texttt{seen})
is the running union of the 8-neighbourhood of the player's cells;
channel 5 (\texttt{enemy\_seen}) the running union of the
8-neighbourhood of visible opponent cells; channels 6--8 the running
unions of visible generals, castles, and mountains. Channel 20 holds
the opponent army count last observed at a cell and keeps it while the
cell is fogged; channel 21 holds $\log(1+\Delta t)/5$ with $\Delta t$
the steps since that observation, reset to $0$ whenever the cell is
visible. Channels 24--37 hold the per-cell differences of own and
opponent army counts over the last seven steps. The two temporal
tokens are sliding windows of the last $512$ values of the opponent's
army and land totals.

\paragraph{Padding and variable map sizes}
Training maps are $17$--$23$ cells on a side and every observation is
padded to $24 \times 24$. Padding cells are marked as mountains once
they have been within sight and as structures in fog before that, so
the board edge behaves like any other terrain the player has or has
not yet seen; scalar channels are broadcast into the padding.

\paragraph{Action masking}
The environment provides a validity mask over source cell and
direction: a move is valid when the source is owned by the player,
holds at least two units, and the destination is on the board and not
a mountain. The mask is applied to the policy logits as a large
negative offset, identically for the all-units and the half-units move
type; passing is always allowed. The value token attends to the
unmasked board.

\section{Top-Advantage Filtering: Per-Seed Results}
\label{app:topadv_sample}

\begin{figure}[t]
    \centering
    \includegraphics[width=\linewidth]{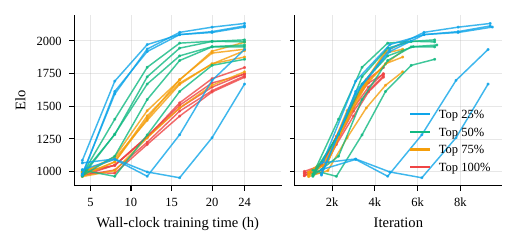}
    \caption{Every run of the filtering ablation (five seeds per
    setting): the checkpoints taken every four hours and the final
    one, against wall-clock time (left) and iterations (right),
    from the same round-robin as Fig.~8.}
    \label{fig:topadv_sample}
\end{figure}

Table~\ref{tab:topadv_seeds} lists every run of the filtering
ablation of Sec.~VII-B and Fig.~\ref{fig:topadv_sample} shows the
runs individually, from the same wall-clock-aligned round-robin as
Fig.~8. After $24$ hours the unfiltered runs end at $1{,}748 \pm 29$
Elo (mean $\pm$ standard deviation over seeds) and the filtered
settings $150$ to $250$ Elo higher. The difference is significant for
$50\%$ and $75\%$ (one-sided Wilcoxon rank-sum test over seeds,
$p = 0.004$ and $0.008$) and for the fifteen filtered runs together
($p = 0.002$); $25\%$ alone reaches $p = 0.08$, as one of its seeds
ends below the unfiltered runs. The three filtered settings are within
noise of one another. Every iteration consumes $262{,}144$ environment frames per
device ($512$ parallel environments $\times$ $512$ turns), yielding
$524{,}288$ player transitions from both seats before filtering.
These counts are independent of the retained fraction, so the
iteration axis is proportional to frames. The equal-frame block of
the table takes each run's checkpoint nearest to iteration $4{,}400$, the last
iteration the unfiltered runs reach ($4{,}400$ for $100\%$, $4{,}500$
for $75\%$, $4{,}500$--$4{,}600$ for $50\%$, $4{,}600$--$4{,}700$
for $25\%$). At equal frames $50\%$ and $75\%$ are ahead of $100\%$
by about $100$ and $80$ Elo (within two standard errors of the
difference), and $25\%$ splits into three seeds about $200$ Elo
ahead and two, seeds $45$ and $46$, still on the early plateau at
that iteration; they leave it after about $20$ and $16$ hours. A $25\%$ iteration takes about $9$ seconds of
wall-clock time and a $100\%$ iteration about $20$, because the
optimizer takes $128$ steps per iteration on the kept quarter and
$512$ on the full batch; this is where the wall-clock gain comes
from. The $25\%$ setting is the most variable across seeds: two
seeds start late, and one of them still ends below every unfiltered
run, while three seeds end above every other run of the sweep.

\begin{table}[t]
    \centering
    \caption{Filtering ablation, every run: tournament Elo of the
    final checkpoint after $24$ hours and of the checkpoint nearest
    to iteration $4{,}400$ (equal frames), both from the round-robin
    of Fig.~8, with the mean $\pm$ standard error over the five seeds
    and the number of iterations completed in $24$ hours.}
    \label{tab:topadv_seeds}
    \footnotesize
    \setlength{\tabcolsep}{4pt}
    \begin{tabular}{@{}lrrrrrcr@{}}
        \toprule
        \textbf{Fraction} & \textbf{44} & \textbf{45} & \textbf{46} & \textbf{47} & \textbf{48} & \textbf{Mean $\pm$ SE} & \textbf{Iter.} \\
        \midrule
        \multicolumn{8}{@{}l}{\textit{Final checkpoint ($24$ h)}} \\
        $25\%$  & 2105 & 1669 & 1934 & 2133 & 2114 & $1991 \pm 88$ & 9448 \\
        $50\%$  & 1954 & 1968 & 1859 & 2007 & 1992 & $1956 \pm 26$ & 6870 \\
        $75\%$  & 1989 & 1763 & 1876 & 1925 & 1935 & $1898 \pm 38$ & 5401 \\
        $100\%$ & 1721 & 1796 & 1731 & 1749 & 1744 & $1748 \pm 13$ & 4437 \\
        \midrule
        \multicolumn{8}{@{}l}{\textit{Iteration $\approx 4{,}400$ (equal frames)}} \\
        $25\%$  & 1971 & 996 & 963 & 1937 & 1918 & $1557 \pm 236$ & \\
        $50\%$  & 1849 & 1885 & 1612 & 1944 & 1982 & $1855 \pm 65$ & \\
        $75\%$  & 1920 & 1660 & 1825 & 1823 & 1906 & $1827 \pm 46$ & \\
        $100\%$ & 1721 & 1796 & 1731 & 1749 & 1744 & $1748 \pm 13$ & \\
        \bottomrule
    \end{tabular}
\end{table}

\section{Network Heads}
\label{app:network}

A single learned \emph{value token} is prepended to every observation
alongside the spatial patch tokens and the two temporal history
tokens. The value head is a linear projection of this value token
followed by a softmax over HL-Gauss bins. The policy head takes only
the tokens corresponding to cells of the game grid and linearly
projects each into a $9$-logit vector, yielding the $H \times W
\times 9$ action distribution.

\section{Hyperparameters}
\label{app:hyperparams}

Table~\ref{tab:hparams} lists the training hyperparameters used in
the reported runs.

\begin{table*}[t]
    \centering
    \caption{Training hyperparameters.}
    \label{tab:hparams}
    \rowcolors{2}{white}{gray!8}
    \begin{tabular}{@{}p{0.28\textwidth}p{0.68\textwidth}@{}}
        \toprule
        \textbf{Parameter} & \textbf{Value} \\
        \midrule
        \multicolumn{2}{@{}l}{\textit{Data}} \\
        Parallel environments per device & $512$ \\
        Rollout length               & $512$ turns \\
        Transitions per iteration    & devices $\times\, 2 \times 512 \times 512$ (both players) \\
        Kept by top-advantage filter & $25\%$ by $|\hat{A}|$ after global normalization; $131{,}072$ per device \\
        Game truncation              & $2{,}048$ turns, reward $0$ \\
        Seeds                        & $44$ (final run, EMA and schedule runs); $44$--$48$ (filter ablation); $44$--$47$ (reward ablation) \\
        Curriculum (BFS spawn distance) & $2$--$6$, $4$--$8$, $6$--$13$, $11$--$17$, $17$--$28$; advance at $\geq 60\%$ wins vs.\ an opponent playing random moves \\
        \midrule
        \multicolumn{2}{@{}l}{\textit{PPO}} \\
        Minibatch size               & $1024$ (so $128$ optimizer steps per iteration) \\
        Epochs per iteration         & $1$ \\
        Clip range $\epsilon$        & $0.2$ \\
        Target KL                    & $0.02$, checked after each epoch (never triggers with one epoch) \\
        Value-loss coefficient $\beta$ & $0.5$ \\
        Gradient-norm clip           & $0.267$ \\
        Optimizer                    & Adam, $\beta_1 = 0.9$, $\beta_2 = 0.999$, $\epsilon = 10^{-8}$, no weight decay \\
        Discount $\gamma$            & $1.0$ \\
        GAE $\lambda$                & $0.9$ \\
        \midrule
        \multicolumn{2}{@{}l}{\textit{Schedules ($t$ = training iteration, $s$ = optimizer step)}} \\
        Learning rate & $\mathrm{clip}\!\left(0.5\,(s/512+1)^{-1.1},\, 5\!\times\!10^{-6},\, 1\!\times\!10^{-4}\right)$ \\
        Entropy coefficient $\alpha$ & $\max\!\left(0.05\,(t+1)^{-0.2},\, 10^{-3}\right)$ \\
        \midrule
        \multicolumn{2}{@{}l}{\textit{Value head (HL-Gauss)}} \\
        Bins                         & $128$ \\
        Range                        & $[-1,1]$ \\
        $\sigma$                     & $0.04$ \\
        \midrule
        EMA decay $\tau$             & $0.999$, updated once per iteration over all parameters \\
        \bottomrule
    \end{tabular}
\end{table*}

\section{Simulator Throughput}
\label{app:sim-throughput}

This section gives the measurements behind Fig.~4 of the main paper and
the protocol used to obtain them. The benchmark scripts, SLURM job files,
and raw results are released with the environment under
\texttt{paper/benchmarks/}.

\paragraph{Definitions}
One \emph{frame} is one environment advanced by one game turn with every
player acting once; a batched call that advances $N$ environments counts
as $N$ frames. A step is the full environment step: the game transition,
the auto-reset of finished games from a pre-generated pool of boards, and
the fog-of-war observations of every player. Both simulators receive uniformly random
actions (a random cell, direction, and split, with a $10\%$ chance of
passing) so that the work per frame is the same; maps are $24 \times 24$;
games are truncated at $500$ turns. The JAX benchmark keeps the
observations of every step in the state of its loop, so the compiler
cannot remove their computation, and the games continue across timed
calls. Compilation time is excluded by a
warm-up call, and each configuration is timed for at least $12$ seconds.
For the NumPy simulator one worker process runs exactly one environment,
so ``parallel environments'' is the same axis for both simulators.

\paragraph{Hardware and software}
Table~\ref{tab:sim-hw} lists the two machines. The CPU node was allocated
whole (all of its memory and $64$ hardware threads) so that no other job
shared it; on shared nodes the same measurements varied by $10$--$80\%$
between runs, which is why the CPU figures are medians of three
repetitions. The prior simulator is the NumPy/Numba implementation of
Straka and Schmid~\cite{straka2024generals} at the commit used for that
paper's own benchmark, driven through a vectorized \texttt{gymnasium}
wrapper with one process per environment.

\begin{table}[H]
    \centering
    \caption{Machines used for the throughput measurements.}
    \label{tab:sim-hw}
    \rowcolors{2}{white}{gray!8}
    \small
    \begin{tabular}{@{}p{0.22\linewidth}p{\dimexpr0.78\linewidth-2\tabcolsep\relax}@{}}
        \toprule
        \textbf{CPU node} & AMD EPYC 7543 (2 sockets $\times$ 32 cores $\times$ 2 threads), 64 threads allocated, whole node \\
        \textbf{GPU} & one NVIDIA H200 NVL (141~GB), driver 575.51, 16 CPU threads allocated \\
        \textbf{JAX simulator} & released commit, JAX~0.11.1, Python~3.12 \\
        \textbf{NumPy simulator} & \texttt{generals-bots} at commit \texttt{b1636da} (April 2025), NumPy~2.2, Numba~0.61 \\
        \bottomrule
    \end{tabular}
\end{table}

\paragraph{Results}
Table~\ref{tab:sim-sweep} contains every point of Fig.~4; all GPU
points come from one job.
Table~\ref{tab:sim-variants} adds the team and free-for-all modes at
$65{,}536$ environments, measured in the same job as the two-player game.
The NumPy simulator runs one process per thread, up to the $64$ allocated
threads, where its repetitions agree within $8\%$. Oversubscribing it
gave no reliable gain: with $256$ processes on $64$ threads the three
repetitions ranged from $5.7$k to $10.4$k frames per second.

\begin{table*}[t]
    \centering
    \caption{Frames per second against the number of parallel
    environments. CPU rows are medians of three runs on one whole node; the
    GPU row is a single run per point. A dash marks a configuration that was not
    measured: the NumPy simulator runs one process per allocated thread, and
    the JAX CPU sweep stops once throughput saturates; $524{,}288$
    environments exceed the GPU's memory.}
    \label{tab:sim-sweep}
    \rowcolors{2}{white}{gray!8}
    \small
    \begin{tabular}{@{}lrrrrrrrrrrr@{}}
        \toprule
        \textbf{Parallel environments} & $1$ & $16$ & $32$ & $64$ & $256$ & $1{,}024$ & $4{,}096$ & $16{,}384$ & $65{,}536$ & $131{,}072$ & $262{,}144$ \\
        \midrule
        NumPy, CPU node & 1.1k & 5.8k & 7.7k & 8.8k & -- & -- & -- & -- & -- & -- & -- \\
        JAX, same CPU node & 8.5k & 48k & -- & 84k & 155k & 411k & 474k & 548k & 684k & -- & -- \\
        JAX, one H200 GPU & 16k & 198k & -- & 757k & 2.5M & 8.8M & 17.9M & 31.4M & 38.1M & 40.8M & 42.8M \\
        \bottomrule
    \end{tabular}
\end{table*}

\begin{table}[H]
    \centering
    \caption{Game modes at $65{,}536$ parallel environments on one H200
    (frames per second).}
    \label{tab:sim-variants}
    \rowcolors{2}{white}{gray!8}
    \small
    \begin{tabular}{@{}lr@{}}
        \toprule
        \textbf{Mode} & \textbf{Frames per second} \\
        \midrule
        1v1 & 38.1M \\
        2v2 & 18.6M \\
        Four-player free-for-all & 18.6M \\
        \bottomrule
    \end{tabular}
\end{table}

\section{Simulator Correctness Tests}
\label{app:sim-tests}

The released environment ships with a test suite of $145$ unit and
property tests (\texttt{pytest}, all passing at the released commit).
Table~\ref{tab:sim-tests} maps the rules of the game to the tests that
exercise them. Most tests are written against small hand-built boards
whose expected outcomes are computed by hand; the mutual-capture tests
additionally sweep $192$ combinations of defending army, attacking stack, split, and turn
against an independent Python oracle. Beyond the unit tests, the agent of
the main paper was trained exclusively in this simulator and played
more than a thousand rated games on the public server through the same
observation pipeline, which would have exposed any discrepancy in the
rules that matters for play.

\begin{table*}[t]
    \centering
    \caption{Rules of \generals{} and the tests that cover them.}
    \label{tab:sim-tests}
    \rowcolors{2}{white}{gray!8}
    \small
    \begin{tabular}{@{}p{0.26\textwidth}p{0.70\textwidth}@{}}
        \toprule
        \textbf{Rule} & \textbf{Tests} \\
        \midrule
        Simultaneous moves and resolution order & \texttt{test\_move\_order}: contested neutral cell, head-on attacks, interception from a third cell \\
        Combat arithmetic, ties, captures & \texttt{test\_game\_jax}, \texttt{test\_mutual\_capture} (192-case oracle sweep; ties hold the cell), \texttt{test\_rules\_coverage} (tied attacks on an enemy cell and on a neutral castle; half move sends $\lfloor n/2 \rfloor$, all-in leaves one), \texttt{test\_multiplayer} (capture requires a strictly larger force) \\
        Move priority (the game's current rule) & \texttt{test\_move\_order}: defensive moves first, attacks on a general last, larger army first, a chased piece after its chaser \\
        Mutual capture in one turn & \texttt{test\_mutual\_capture}, \texttt{test\_deathtouch}: reported as a draw under the competition rules \\
        Army production and scoreboard & \texttt{test\_game\_jax}, \texttt{test\_build\_castles}, \texttt{test\_rules\_coverage}: generals and castles every second turn, phase of the first increment, every owned plain cell on the 50-turn tick only, land and army totals exact through capture and production \\
        Fog of war and visibility & \texttt{test\_game\_jax}, \texttt{test\_rules\_coverage} (visible set is exactly the owned cells plus their 8-neighbourhood; \texttt{structures\_in\_fog} exact), \texttt{test\_multiplayer} (team-shared sight) \\
        Map generation and spawn distance & \texttt{test\_spawn\_distance}, \texttt{test\_grid\_generation\_performance}, \texttt{test\_rules\_coverage}: BFS distance floor, both generals present and connected, mountain fraction within the requested band, castle counts \\
        Team play (2v2) & \texttt{test\_multiplayer}: ally merge transfers ownership, team loses only when all its generals fall, allies share the result, shared visibility \\
        Free-for-all & \texttt{test\_multiplayer}: capture transfers all cells with armies halved, general becomes a castle, victim eliminated and ignored, last player standing wins \\
        Truncation and rewards & \texttt{test\_rules\_coverage}, \texttt{test\_multiplayer}: truncated flag exactly at the limit, zero reward on truncation, auto-reset, $\pm 1$ by team result \\
        Optional rule modifiers & \texttt{test\_build\_castles}, \texttt{test\_deathtouch} \\
        \bottomrule
    \end{tabular}
\end{table*}

\paragraph{Agreement with recorded games}
We validate the simulator against the version of \generals{} live on
September 20, 2026: $10{,}000$ ranked two-player games were replayed
through the simulator and compared with the game's own engine. A replay stores the starting board (mountains, castles with
their starting units, the two generals) and the ordered list of moves,
each with its player, source and destination cell, a half-move flag,
and the turn on which it was issued; games that end by a resignation
also carry that event. The games were drawn at random from the ranked
games played between the start of the deployment window of Sec.~VI and
September 20, 2026 (the two-month window itself and $1{,}000$ games of
the final day), all with at least $50$ turns. Two event weekends on which the ladder ran optional map
features (lookouts and tunnels, $396$ games in the pool) are not part
of the base game and were excluded before sampling.

The reference is the game's own engine: the JavaScript the site ships
to play replays back in the browser, which implements the current
rules, run offline on the same replay files to record the board after
every turn. A game \emph{replays perfectly} when three conditions hold:
(i) after every turn from the first to the last, the simulated board
agrees with the reference on the owner and the army count of every
cell; (ii) no recorded move is rejected by the simulator, that is, at
the start of its turn the mover owns the source cell with at least two
armies; and (iii) the game ends as recorded, by the recorded capture on
the recorded turn with the recorded winner, or, for a game ended by a
resignation, not before it. Condition (i) is the strong one: a wrong
production phase, combat rule, capture rule, or resolution order shows
up as a differing cell within a few turns and then compounds.

All $10{,}000$ games replay perfectly (Table~\ref{tab:replay}). The
only difference is the board after the final capture, where the game's
engine still applies that turn's army income to the winner while the
simulator stops at the capture; it affects no game. The same check on
the other modes gives the same result: $1{,}000$ ranked free-for-all
games (up to nine players) and $1{,}000$ ranked 2v2 games, drawn
uniformly from the month before September 21, 2026, all replay
perfectly, with the game's server-side events (a player leaving, and the
later hand-over of that player's land) applied as recorded. The replays, the scripts, the recorded reference boards, and the
per-game results are released under \texttt{paper/validation/}; the
whole check runs in about a quarter of an hour on a laptop.

\begin{table}[H]
    \centering
    \caption{Replaying $10{,}000$ ranked games, drawn from the
    deployment window until September 20, 2026, through the simulator,
    compared with the game's own engine.}
    \label{tab:replay}
    \rowcolors{2}{white}{gray!8}
    \small
    \begin{tabular}{@{}p{0.72\columnwidth}r@{}}
        \toprule
        Games (random sample, current rules) & $10{,}000$ \\
        Recorded moves & $6{,}306{,}208$ \\
        Turns compared & $3{,}884{,}961$ \\
        Games whose board is identical to the game's engine after every turn & $10{,}000$ \\
        Recorded moves rejected by the simulator & $0$ \\
        Games ending by capture: recorded winner on the recorded turn & $6{,}236$ of $6{,}236$ \\
        Games ending by resignation: no earlier end & $3{,}764$ of $3{,}764$ \\
        \bottomrule
    \end{tabular}
\end{table}

\section{Ablation Details}
\label{app:ablation-details}

\paragraph{Local Elo}
Each local Elo figure comes from one round-robin among the checkpoints
it compares. Every pair of checkpoints plays an equal number of games
in each seat on boards drawn from the full-game distribution
($17$--$23$ boards with the generals at least $17$ steps apart; the
filter ablation of Fig.~8 and the reward ablation of Fig.~S.2 use the
$18 \times 18$ boards they were trained on). A single Elo fit per figure, initialized at $1500$ for
every checkpoint and optimized by $100$ shuffled passes of stochastic
updates with $k = 0.01$, places them on one scale. Fig.~S.2 has $256$ games per pair among $104$ checkpoints;
Fig.~8 has
$256$ games per pair among $120$ checkpoints (each run's checkpoints at $4, 8, \ldots, 20$ hours, the saved
checkpoint nearest to the hour, and its final one); Fig.~9
(right) has $250$ to $500$ games per pair. Differences
within one figure are the meaningful quantity; values are not
comparable across figures. The EMA parameters are the ones evaluated
unless stated otherwise.

\paragraph{EMA versus last iterate}
Fig.~7 uses direct games. The three large-scale runs, each trained on
four H200 GPUs for $39{,}000$ iterations, save the EMA (decay $0.999$)
and the last-iterate parameters at the same iterations.
At every $1{,}000$ iterations from $1{,}000$ to $39{,}000$ of each run, the
EMA checkpoint plays the last-iterate checkpoint of the same iteration,
$250$ games in each seat, on the full-game boards described above, with
one fixed key per iteration so that the three runs play the same boards
at the same iteration; $117$ pairs and $58{,}500$ games in all. The lead
is $400 \log_{10}\!\big(s / (1 - s)\big)$ for the EMA's score $s$, with a
draw counting half; the figure shows its mean over the three runs and
the range across them.

\paragraph{Evaluation during training}
During training the current parameters play $512$ games greedily
against an opponent that plays uniformly random moves, half in each
seat; the win rate drives the curriculum (Sec.~V-C). The EMA and the
last-iterate parameters are saved every $500$ iterations.

\paragraph{Reward shaping}
The eight runs of Sec.~VII-C share one configuration and differ only
in the reward: the sparse $\pm 1$ at the end of the game, or the same
reward plus $\Phi(s') - \Phi(s)$ with $\gamma = 1$ and
\begin{equation*}
\Phi(s) = 0.3\,\mathrm{clip}\!\left(\tfrac{\log(a/a')}{\log 1.6}, -1, 1\right)
        + 0.3\,\mathrm{clip}\!\left(\tfrac{\log(l/l')}{\log 1.3}, -1, 1\right),
\end{equation*}
where $a, a'$ and $l, l'$ are the two players' army and land totals.
The terminal potential is zero, so the shaping term telescopes to a
constant of the start state and the optimal policy is
unchanged~\cite{ng1999pbrs}. The shared configuration is a transformer
with embedding $384$, depth $5$, and
feed-forward factor $3$; $18 \times 18$ boards with the spawn-distance
curriculum of Sec.~V-C; $512$ environments and $512$ steps per
iteration with every transition kept; seeds $44$ to $47$ per reward;
$24$ hours on one H200 each, about $5{,}000$ iterations, then resumed
from the iteration-$5{,}000$ checkpoint with its optimizer state and
schedules for a second $24$ hours, to about $10{,}000$. The shaped
runs add the potential-based term on every curriculum
stage. The EMA checkpoints every $500$ iterations to $5{,}000$ and at
$6{,}000$, $8{,}000$, and $10{,}000$, $104$ in total, play one
round-robin with $256$ games per pair on $18 \times 18$ boards with
the generals $14$--$24$ steps apart, truncated at $1{,}024$ turns,
with the Elo fit described above.

\begin{figure}[t]
    \centering
    \includegraphics[width=\linewidth]{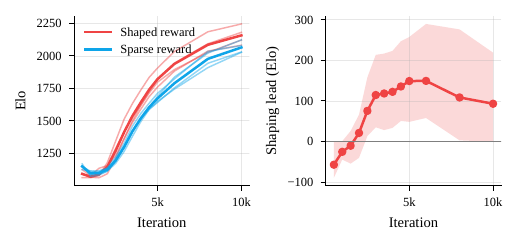}
    \caption{Reward-shaping ablation. \textbf{Left:} local Elo of runs
    trained with shaped reward (red) and sparse win/loss reward (blue)
    over $10{,}000$ iterations, four seeds each (thin) and their mean
    (thick). \textbf{Right:} Elo lead of each shaped run over the sparse
    run with the same seed at the same iteration: mean over the four
    seed pairs (line) and the range across them (band).}
    \label{fig:reward}
\end{figure}

Fig.~\ref{fig:reward} shows the result. Shaping is behind on average
for the first $1{,}500$ iterations and ahead in every seed pair from
$2{,}500$ to $6{,}000$ iterations, by $149$ Elo on average at $5{,}000$
and $6{,}000$; at $5{,}000$ the shaped runs win $75\%$ of direct games
against the sparse runs of the same iteration. By $10{,}000$ the mean
lead is $93$ Elo ($65\%$ of direct games), and it depends on the
seed: two pairs are level ($+2$ and $-1$ Elo) and two keep a large
lead ($+152$ and $+219$).

\paragraph{Compute}
Table~\ref{tab:compute} lists the training compute behind every
experiment. Training jobs ran on NVIDIA H200 NVL GPUs.

\begin{table}[t]
    \centering
    \caption{Training compute per experiment. GPU-hours are wall-clock
    hours times GPUs allocated.}
    \label{tab:compute}
    \scriptsize
    \setlength{\tabcolsep}{3pt}
    \begin{tabular}{@{}lrlr@{}}
        \toprule
        \textbf{Experiment} & \textbf{Runs} & \textbf{Per run} & \textbf{GPU-h} \\
        \midrule
        Final agent (Sec.~VI) & 1 & $4\times$H200, $7$ days & $672$ \\
        Schedule runs (Fig.~9) & 3 & $4\times$H200, $7$ days & $2{,}016$ \\
        Filter ablation (Fig.~8) & 20 & $1\times$H200, 24 h & 480 \\
        Reward ablation (Fig.~S.2) & 8 & $1\times$H200, 48 h & 382 \\
        \bottomrule
    \end{tabular}
\end{table}

\section{Ladder Evaluation}
\label{sec:supp-ladder}
This section documents the fixed-checkpoint match evaluation of Sec.~VI from the
platform's own records. The game lists of both accounts were
retrieved from the public replay index of \generals{}, which stores
for every game its identifier, start time, both players' account
names, and both players' ratings at that time. So that the record does
not depend on the platform keeping it, the release contains the replay
file of each of the $1{,}147$ rated games of the window together with
this metadata, a script that replays any of them in the simulator's
viewer, and the per-game results of replaying them (all $1{,}136$ games
with moves replay perfectly, Sec.~S-VI); the official season standings
pages were also saved to the Internet Archive.

\paragraph{Accounts and window}
The deployed checkpoint is the EMA parameters of the final run
(\texttt{L\_7d\_gae90}) at iteration $38{,}000$, the same on both
accounts for the whole window. It played under \texttt{Average Joe} and
\texttt{L\_7d\_gae90\_30k\_ema} (two accounts to gather games at
double rate). The first account had been used for
earlier development agents before the window; those games are not
part of the evaluation. The window is the two months from
2026-04-08, the first day of the second account, to 2026-06-08
inclusive; it overlaps ten of the site's standings weeks, season~43
week~7 to season~44 week~6. It holds $1{,}147$ rated 1v1 games against
$173$ accounts, $596$ and $551$ on the two accounts; $31$ rated games
between our two accounts and $67$ unrated custom games in the same
period are excluded, except the $23$-game custom series against
\texttt{zero v3} in Table~III of the main paper; its one loss, replay
\texttt{C-sINIgzc}, was a connection failure on our side. No games
were pre-arranged.
The low displayed ratings after gaps do not by themselves establish
an inactivity reset. The global replay record supports a season
reset on May~3; account-state and inactivity reconstruction are
examined separately in Sec.~\ref{sec:supp-reconstruction}.

\paragraph{Opponents}
Table~\ref{tab:ladder-opponents} lists every opponent with at least
ten games in the window, with the accounts that the two top players
confirmed as theirs merged into one row. Human and bot opponents are
marked. The rating intervals of Table~III of the main paper count human opponents only.

\begin{table}[t]
    \centering
    \caption{Opponents with at least ten rated games in the window.
    Peak is the opponent's highest rating at the time of a game
    against the agent. \texttt{shimatetsu} includes \texttt{Reisen};
    \texttt{Mithraaaa} includes \texttt{Sinyth} and \texttt{Perihelion}.}
    \label{tab:ladder-opponents}
    \footnotesize
    \begin{tabular}{lrrr}
        \toprule
        Opponent & W & L & Peak \\
        \midrule
        \texttt{Mithraaaa} & 184 & 65 & 106 \\
        \texttt{Kongbai} & 112 & 18 & 95 \\
        \texttt{shimatetsu} & 25 & 17 & 109 \\
        \texttt{hkjkgh} & 37 & 5 & 97 \\
        \texttt{peacemaker 2} (bot) & 32 & 7 & 85 \\
        \texttt{satsukino} & 26 & 6 & 100 \\
        \texttt{0dd} & 22 & 9 & 96 \\
        \texttt{eVen} & 18 & 6 & 93 \\
        \texttt{KyIe VaIentine} & 12 & 9 & 100 \\
        \texttt{rubbish server} & 13 & 2 & 91 \\
        \texttt{GSW NBA Champion} & 13 & 2 & 89 \\
        \texttt{asdvlaksudhiouwenf} & 11 & 4 & 86 \\
        \texttt{Hammer U} & 10 & 4 & 94 \\
        \texttt{AtomicGamer2109} & 14 & 0 & 53 \\
        \texttt{Syn28-19306} & 10 & 4 & 84 \\
        \texttt{libai} & 12 & 2 & 89 \\
        \texttt{Human.exe} (bot) & 13 & 0 & 77 \\
        \texttt{zzd\_bot release} (bot) & 13 & 0 & 71 \\
        \texttt{Alexithymia} & 12 & 1 & 87 \\
        \texttt{green321} & 10 & 1 & 86 \\
        \texttt{Four side attack} & 10 & 0 & 86 \\
        \texttt{GoldenKela} & 8 & 2 & 95 \\
        \texttt{EklipZ} & 9 & 1 & 75 \\
        \texttt{I Got Smoke} & 6 & 4 & 82 \\
        \bottomrule
    \end{tabular}
\end{table}

\paragraph{Games against the top humans after the window}
Table~III of the main paper extends the records against the two
highest-rated humans to 2026-09-20; the same checkpoint played the
$15$ rated games after the window. Against \texttt{Mithraaaa}, all as
\texttt{Sinyth} ($5$--$2$): 2026-06-10 \texttt{sQWsl1OMkH}~(W, second
account); 2026-06-10 \texttt{D9XkFJB0g}~(W); 2026-06-22
\texttt{-yPJRgASM}~(L, second account); 2026-09-15
\texttt{y76mnthyF}~(W); 2026-09-20 \texttt{ZXPTQxmsc}~(L); 2026-09-20
\texttt{OAri6df-m}~(W); 2026-09-20 \texttt{TAAKmA-Ix}~(W). Against
\texttt{shimatetsu}, all as \texttt{Reisen} ($7$--$1$): 2026-09-14
\texttt{ABEnj8LS1}~(W); 2026-09-14 \texttt{mXUwqFPl3}~(W); 2026-09-20
\texttt{6xR4pJgSg}~(W); 2026-09-20 \texttt{EOBDgTspL}~(W); 2026-09-20
\texttt{Ky95aMrVS}~(W); 2026-09-20 \texttt{Lw8fdccwQ}~(L); 2026-09-20
\texttt{1da\_pARyu}~(W); 2026-09-20 \texttt{BK95B\_hxy}~(W). Games
without a note were played on \texttt{Average Joe}.

\begin{figure}[t]
    \centering
    \includegraphics[width=\linewidth]{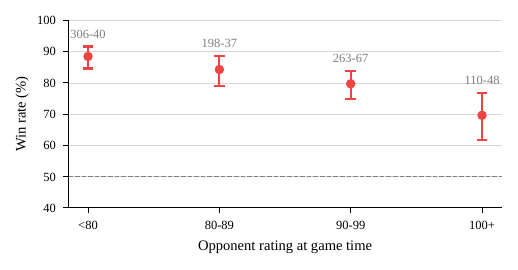}
    \caption{Win rate against human opponents by their rating at
    game time, with $95\%$ Clopper--Pearson intervals; the dashed
    line marks $50\%$.}
    \label{fig:ladder-bands}
\end{figure}

\paragraph{Win rate by opponent rating}
Fig.~\ref{fig:ladder-bands} plots the win rate against human
opponents by their rating at game time with $95\%$ Clopper--Pearson
intervals; the values are those of Table~III of the main paper. A logistic regression
of the game outcome on the opponent's rating, over the $1{,}069$
games against humans, gives a win probability of $0.68$ against a
$109$-rated opponent ($95\%$ bootstrap interval over games
$0.62$--$0.74$) and places the opponent rating at which the agent
would win half of its games at $132$ ($120$--$151$). Fitted on the
$488$ games against opponents rated $90$ or above only, the same
quantity is $123$ with a wide interval ($108$--$217$). The platform
does not expose the uncertainty of its ratings, so we report the
outcomes directly.

\paragraph{Deployment and latency}
The agent ran on a laptop with an AMD Ryzen~5 PRO 6650U CPU and no
GPU, connected to the game server through the public bot API, and
issued one action per $0.5$-second tick. Over $5{,}190$ recorded
turns of $23$ games, inference took $32$~ms per turn (median), $37$~ms
at the 99th percentile, and at most $45$~ms after the first turn of
a game.
In a separate offline CPU check of the deployed checkpoint, peak
resident memory stayed below $2$~GiB, including model loading and
runtime overhead. The check used the same CPU model and ran $1{,}980$
inference calls over all $49$ board shapes with each side from $17$
to $23$ cells, with rendering and debug recording disabled.

\section{Official Weekly Standings and Rating Reconstruction}
\label{sec:supp-reconstruction}

This analysis complements the fixed-checkpoint match evaluation in
Sec.~\ref{sec:supp-ladder}. It asks whether the deployed accounts'
first-place standing persists when all available ranked games are
replayed with common initial ratings and confirmed alternate
accounts are combined. The reconstruction does not recover the
site's hidden rating states; it is a separate rating of the same games.

\subsection{Ten Official Weekly Snapshots}
We use the platform's archived 1v1 standings for the ten weeks that
overlap the two-month deployment, season~43, weeks~7--10, and
season~44, weeks~1--6; the first and last of these weeks are only
partly inside the window.\footnote{Official archives:
\url{https://generals.io/rankings/43} and
\url{https://generals.io/rankings/44}.}
An agent account is first in each of these ten consecutive
end-of-week snapshots (Table~\ref{tab:official-weeks}). We compare
the accounts as the website displayed them, without merging their
official stars. The snapshots are one longitudinal record.

\begin{table}[t]
    \centering
    \caption{Official weekly standings. A: \texttt{Average Joe};
    B: \texttt{L\_7d\_gae90\_30k\_ema}. The agent is rank~1 in every
    row. Other is the highest account excluding our two accounts
    and the three designated bots. Values are official stars.}
    \label{tab:official-weeks}
    \footnotesize
    \setlength{\tabcolsep}{3pt}
    \begin{tabular}{@{}ccrlrr@{}}
        \toprule
        Season/week & Agent & Stars & Other account & Stars & Gap \\
        \midrule
        43/7 & B & 109.91 & shimatetsu & 107.26 & 2.65 \\
        43/8 & B & 112.35 & shimatetsu & 102.31 & 10.04 \\
        43/9 & A & 114.13 & shimatetsu & 107.83 & 6.30 \\
        43/10 & A & 114.33 & Perihelion & 107.85 & 6.48 \\
        44/1 & A & 103.55 & Perihelion & 97.90 & 5.65 \\
        44/2 & B & 111.80 & Perihelion & 98.51 & 13.29 \\
        44/3 & B & 105.38 & Sinyth & 100.01 & 5.37 \\
        44/4 & B & 117.07 & zZz & 98.85 & 18.22 \\
        44/5 & B & 119.48 & I Got Smoke & 99.05 & 20.43 \\
        44/6 & A & 121.92 & d3st1ny & 105.15 & 16.77 \\
        \bottomrule
    \end{tabular}
\end{table}

\subsection{Collection, Timing and Identity Resolution}
We collect the public global replay index with overlapping pages of
at most $200$ records, retain the raw JSON, and deduplicate by replay
identifier. Two collections padded around the window yield
$359{,}721$ records across game modes, including $202{,}585$ ranked
1v1 records. All adjacent-page overlap checks pass, and every game in
the overlap with the $65$ previously cached account histories agrees
in timestamp, participant names, outcome and recorded stars.

We keep records with type \texttt{1v1} and ladder \texttt{duel}.
The API supplies game starts and turn counts, but not exact rating
update times. Games are sorted by estimated finish,
\begin{equation}
t_{\mathrm{finish}}=t_{\mathrm{start}}
             +0.5\,\max(\mathrm{turns}-1,0)\ \mathrm{s},
\end{equation}
breaking ties by start time and replay identifier. The window runs
from 2026-04-08 00:00 UTC to 2026-06-08 23:59 UTC by estimated
finish. The site's season reset on May~3 falls inside it and does
not affect the recomputation, which starts every player from a
common rating.

We resolve account renames with the API's \texttt{currentName} where
available, avoiding conflation of reused historical names. We then
apply the confirmed ownership groups in Table~\ref{tab:rating-aliases}.
Human account ownership is based on player or author confirmation,
not similarity of names or match outcomes. Each group has one
initial state and all its games update that state. Pooling assumes one
player behind the merged accounts and can change both the mean and
the uncertainty; unknown alternate accounts remain separate.

\begin{table*}[t]
    \centering
    \caption{Confirmed identity groups applied before recomputing
    Elo and OpenSkill. Some listed aliases have no games in the
    window; their inclusion does not add observations.}
    \label{tab:rating-aliases}
    \small
    \begin{tabular}{@{}p{0.17\textwidth}p{0.79\textwidth}@{}}
        \toprule
        Identity & Accounts \\
        \midrule
        Agent & \texttt{Average Joe}, \texttt{L\_7d\_gae90\_30k\_ema} \\
        shimatetsu & \texttt{shimatetsu}, \texttt{Reisen} \\
        Mithraaaa & \texttt{Mithraaaa}, \texttt{Sinyth}, \texttt{Perihelion} \\
        I Got Smoke & \texttt{szmD}, \texttt{I Got Smoke}, \texttt{O=O}, \texttt{Pi Day}, \texttt{safhubsabhkj}, \texttt{Love in a Puff} \\
        \bottomrule
    \end{tabular}
\end{table*}

The finish-time window contains $173{,}815$ ranked records. We
exclude eight malformed records whose two positions resolve to the
same account and $31$ games between our own merged accounts. All
other games remain, including short games and games with no kills.
The resulting dataset contains $173{,}776$ games and $14{,}616$
active merged identities.

Our conservative human comparison pool excludes the agent and the
three bot identities designated in the reconstruction protocol:
\texttt{Human.exe}, \texttt{bike9800} and \texttt{zzd\_bot release}.
The remaining $14{,}613$ identities may include further bots, so the
pool is a superset of the human field. Participation filters below
only restrict who can be the strongest comparator; they never remove
games or refit the ratings.

\subsection{Chronological Elo and Day Bootstrap}
Every identity starts at $1500$. For a game between $i$ and $j$,
the expected score and update are
\begin{align}
e_i&=\left(1+10^{(R_j-R_i)/400}\right)^{-1},\\
R_i'&=R_i+K(y_i-e_i),\qquad
R_j'=R_j-K(y_i-e_i),
\end{align}
where $y_i=1$ for a win and $0$ for a loss. We make one
chronological pass at each of $K=10,20,40$, without calendar decay,
and report all three. (The local Elo of the training ablations,
Sec.~\ref{app:ablation-details}, is a different, repeated-pass fit.)

For a fixed comparison set $H$, the statistic of interest is
\begin{equation}
\Delta_K=R_A-\max_{h\in H}R_h,
\end{equation}
where $A$ is the merged agent. We partition games by their UTC
calendar day of estimated completion, giving $D=62$ blocks. In each of $20{,}000$
resamples (seed $2048$), we sample $D$ day blocks with replacement.
Every game in a selected day receives that day's common integer
multiplicity. We sort the resulting multiset by the original game
order, so repeated copies of a game are adjacent, and replay Elo
from $1500$. This preserves chronological ordering while resampling
whole days of outcomes. We recompute the maximum over the entire
fixed set $H$ in every draw, rather than comparing only with the
observed runner-up.

The fifth percentile of the resampled $\Delta_K$ is the nominal
one-sided $95\%$ lower bound for a positive lead; the $2.5$th and
$97.5$th percentiles give a two-sided interval. The $p$-values of
Sec.~VI-C are the fractions of resamples with $\Delta_K \le 0$. The bounds
quantify the variability of this Elo procedure under day resampling;
they do not cover dependence beyond a day, model misspecification, or
the retrospective choice of window and analysis, and a player absent
from a resample keeps the initial rating.

\begin{table}[t]
    \centering
    \caption{Elo recomputed over the two-month window. The agent
    ranks first in all rows. Lead is relative to the strongest identity in the
    conservative human comparison pool; the final column is its
    one-sided $95\%$ day-bootstrap lower bound.}
    \label{tab:ladder-reconstruction}
    \small
    \begin{tabular}{@{}rrrr@{}}
        \toprule
        $K$ & Agent Elo & Lead & Lower bound \\
        \midrule
        10 & 2002.5 & $+154.4$ & $+66.4$ \\
        20 & 2127.5 & $+136.8$ & $-6.6$ \\
        40 & 2305.1 & $+155.6$ & $-101.6$ \\
        \bottomrule
    \end{tabular}
\end{table}

With all $14{,}613$ comparators (Table~\ref{tab:ladder-reconstruction}), the lower bounds are $+66.4$,
$-6.6$ and $-101.6$ Elo at $K=10,20,40$, respectively. At $K=10$
the two-sided interval is $[+49.3,+227.8]$, so the lead is
significant under both criteria; at $K=20$ it is $[-32.9,+216.3]$,
and the Monte Carlo interval of the one-sided bound, about
$[-8.8,-4.3]$ Elo, shows that its sign is not a resampling artifact.
$K$ sets the memory of a rating, about $400/K$ games; $K=10$ is the
value FIDE uses for established players rated $2{,}400$ and
above~\cite{fide2024rating}, and among the ten highest-rated
players the agent's $1{,}147$ games are by far the most (the next has
$403$), whereas $K=40$ is the
provisional-player setting that reflects only the last ten or so
games of every player.

\begin{table*}[t]
    \centering
    \caption{Candidate-only participation sensitivity at $K=20$.
    Every row uses the same complete game network and bootstrap
    draws. These thresholds are retrospective, not independently
    specified confirmation populations.}
    \label{tab:rating-scopes}
    \small
    \begin{tabular}{@{}lrrr@{}}
        \toprule
        Candidates & Count & One-sided $95\%$ lower bound & Two-sided $95\%$ interval \\
        \midrule
        All comparison identities & 14{,}613 & $-6.6$ & $[-32.87,+216.29]$ \\
        $\geq20$ games & 2{,}229 & $-6.6$ & $[-32.87,+216.29]$ \\
        $\geq50$ games & 1{,}055 & $-3.9$ & $[-30.01,+219.23]$ \\
        $\geq100$ games & 571 & $+17.0$ & $[-7.38,+246.82]$ \\
        $\geq100$ games, $5$ opponents, $5$ days & 540 & $+17.0$ & $[-7.38,+246.85]$ \\
        \bottomrule
    \end{tabular}
\end{table*}

At $K=20$ the $571$ players with at least $100$ games give a
one-sided bound of $+17.0$ Elo (Table~\ref{tab:rating-scopes}), a
two-sided interval of $[-7.4,+246.8]$, and a Monte Carlo interval of
$[+14.7,+19.4]$ for the bound; adding at least five opponents and
five active days leaves $540$ players and the same bound. The
$\geq50$-game group ($1{,}055$ players) behaves like the whole field:
the strongest competitor in a resample is often ($43\%$ of
resamples) a player with between $50$ and $100$ games and a good
stretch.
The archive retains all tested participation scopes, including
unsuccessful ones. No activity threshold makes the reported
$K=40$ one-sided bound positive.

\subsection{OpenSkill Reconstruction and Model Uncertainty}
We use the Plackett--Luce OpenSkill update with its default
uncertainty-reduction function~\cite{openskill,weng2011ranking},
initializing every merged identity at $\mu_0=50$, $\sigma_0=50/3$,
with $\beta=50/6$ and
$\tau\in\{0,25/300,50/300\}$. Before a game, the update adds
$\tau^2$ to each participant's variance. We set the numerical
floor to $10^{-4}$ and do not cap sigma or freeze its reduction on
rematches. The implementation agrees with
\texttt{openskill.js} version~4.1.1 to numerical precision.
The agent is first by both mean $\mu$ and conservative ordinal
$\mu-3\sigma$ under every setting.

We export each player's approximate Gaussian marginal interval
$\mu\pm1.96\sigma$. Under independent Gaussian skill marginals,
the pairwise comparison is
\begin{equation}
P(S_A>S_h)\approx
\Phi\!\left(\frac{\mu_A-\mu_h}
{\sqrt{\sigma_A^2+\sigma_h^2}}\right).
\end{equation}
Here $\Phi$ and $\phi$ denote the standard normal cumulative
distribution and density. To compare with the whole field, we integrate
\begin{equation}
P(S_A>\max_{h\in H}S_h)\approx
\int_{-\infty}^{\infty}\!\phi(z)
\prod_{h\in H}\Phi\!\left(\frac{\mu_A+\sigma_Az-\mu_h}{\sigma_h}\right)\,dz.
\end{equation}
We also report a conservative threshold union bound: for any $t$,
\begin{equation}
P(S_A>\max_h S_h)\geq
\left[1-P(S_A\leq t)-\sum_h P(S_h\geq t)\right]_+.
\end{equation}
We optimize $t$ by a grid and local refinement. The bound allows
arbitrary dependence between players but assumes the Gaussian
marginals.

\begin{table*}[t]
    \centering
    \caption{Merged-player OpenSkill at the end of the window. Rank is
    by $\mu$ / $\mu-3\sigma$. $P_{\rm all}$ and $P_{100}$ are joint
    probabilities under independent Gaussian marginals for all
    $14{,}613$ comparators and the $571$ with at least $100$ games;
    $L_{100}$ is the Gaussian marginal union lower bound. All values
    are conditional on the specified model.}
    \label{tab:reconstructed-openskill}
    \small
    \setlength{\tabcolsep}{4.5pt}
    \begin{tabular}{@{}lrrrrrrr@{}}
        \toprule
        Calendar rule & $\tau$ & $\mu_A$ & $\sigma_A$ & Rank & $P_{\rm all}$ & $P_{100}$ & $L_{100}$ \\
        \midrule
        No calendar drift & 0.00000 & 117.222 & 2.049 & 1 / 1 & 16.22\% & 97.70\% & 85.49\% \\
        No calendar drift & 0.08333 & 119.979 & 2.589 & 1 / 1 & 36.04\% & 98.85\% & 89.61\% \\
        No calendar drift & 0.16667 & 124.031 & 3.424 & 1 / 1 & 61.90\% & 99.27\% & 91.71\% \\
        Inactivity sensitivity & 0.00000 & 127.565 & 3.539 & 1 / 1 & 0.00\% & 65.41\% & 17.38\% \\
        Inactivity sensitivity & 0.08333 & 128.695 & 3.727 & 1 / 1 & 0.00\% & 68.09\% & 22.46\% \\
        Inactivity sensitivity & 0.16667 & 130.915 & 4.156 & 1 / 1 & 0.00\% & 72.16\% & 29.96\% \\
        \bottomrule
    \end{tabular}
\end{table*}

\paragraph{Inactivity sensitivity}
The baseline leaves states unchanged between games. The official
changelog describes approximately one star per day of decay after
$48$ hours of inactivity, with an adjustment for ladder availability,
but does not specify a sigma update~\cite{generalsversions}. We test
the unverified rule
\begin{equation}
\mu^- = \mu,\qquad
\sigma^- = \sigma+\frac{\max(0,d-2)}{3},
\end{equation}
where $d$ is wall-clock days since the last game. This reduces
$\mu-3\sigma$ by one per additional inactive day. The adjustment
is applied before the next game and when evaluating an endpoint,
with no inflation before a player's first game and no correction
for ladder closures. Any confirmed alias playing resets the merged
identity's clock. We apply the rule throughout the history, not
only to final display values.

The rule was chosen on the season-44 record because it lowers the
RMSE against the recorded stars from $1.69$ to $1.36$, with the
fraction of later integer star values matched exactly unchanged at
$69\%$; it is not the server's code, and the site's exact sigmas are
not established.

The merged agent remains first under this sensitivity, while joint
probabilities change materially (Table~\ref{tab:reconstructed-openskill}).
For the $571$ comparators with at least $100$ games, baseline
probabilities of $97.7$--$99.3\%$ become $65.4$--$72.2\%$.
The much lower all-identity probabilities reflect the large number
of inactive players whose uncertainty grows, not a reversal of the
agent's point rank. We report both models rather than select the
favorable one.

\subsection{Reproducibility}
The released archive holds the raw index pages, the deduplicated game
export, the identity mapping, excluded records, every player's Elo and
OpenSkill states, the bootstrap samples, and the script that generates
Fig.~6 and the tables of this section from them.

\end{document}